\documentclass[10pt,final,journal,letterpaper,twoside,twocolumn]{IEEEtran}

\usepackage[numbers,sort]{natbib}
\usepackage{multicol}
\usepackage{multirow}
\usepackage[bookmarks=true]{hyperref}
\usepackage{xspace}
\usepackage{multirow}

\usepackage{amsmath}  
\usepackage{amssymb}
\usepackage{mathtools}
\usepackage{color}
\usepackage{booktabs} 
\usepackage{lipsum} 
\usepackage{framed}
\usepackage[pdftex]{graphicx}
\usepackage{rotating}
\usepackage[font=footnotesize]{caption}
\usepackage{subcaption}
\usepackage{prettyref}
\usepackage{pifont}
\usepackage{epstopdf}
\usepackage{xkvltxp}

\usepackage[usenames]{xcolor}
\usepackage{mdframed}
\usepackage{tcolorbox}

\newcommand{\toAdd}[1]{} 

\usepackage{array}
\newcolumntype{L}[1]{>{\raggedright\let\newline\\\arraybackslash\hspace{0pt}}m{#1}}
\newcolumntype{C}[1]{>{\centering\let\newline\\\arraybackslash\hspace{0pt}}m{#1}}
\newcolumntype{R}[1]{>{\raggedleft\let\newline\\\arraybackslash\hspace{0pt}}m{#1}}

\newcommand{\setal}{~\emph{et al.}}

\newcommand{\myParagraph}[1]{{\bf #1.\xspace}}
\newcommand{\mySubParagraph}[1]{\emph{#1:}}

\newcommand{\Real}[1]{\mathbb{R}^{#1}}
\newcommand{\blue}[1]{{\color{black}#1}}

\newcommand{\YL}[1]{\blue{#1}}
\newcommand{\notAddressed}[1]{\xspace} 
\let\labelindent\undefined
\usepackage{enumitem}

\newcommand{\aopt}{\textit{A-opt}}
\newcommand{\dopt}{\textit{D-opt}}
\newcommand{\eopt}{\textit{E-opt}}

\newcommand{\calX}{{\cal X}}

\newcommand{\tran}{^{\mathsf{T}}}

\newrefformat{prob}{Problem\,\ref{#1}}
\newrefformat{def}{Definition\,\ref{#1}}
\newrefformat{sec}{Section\,\ref{#1}}
\newrefformat{sub}{Section\,\ref{#1}}
\newrefformat{prop}{Proposition\,\ref{#1}}
\newrefformat{app}{Appendix\,\ref{#1}}
\newrefformat{alg}{Algorithm\,\ref{#1}}
\newrefformat{cor}{Corollary\,\ref{#1}}
\newrefformat{thm}{Theorem\,\ref{#1}}
\newrefformat{lem}{Lemma\,\ref{#1}}
\newrefformat{fig}{Fig.\,\ref{#1}}
\newrefformat{tab}{Table\,\ref{#1}}

\newcommand{\bdmath}{\begin{dmath}}
\newcommand{\edmath}{\end{dmath}}
\newcommand{\beq}{\begin{equation}}
\newcommand{\eeq}{\end{equation}}
\newcommand{\bdm}{\begin{displaymath}}
\newcommand{\edm}{\end{displaymath}}
\newcommand{\bea}{\begin{eqnarray}}
\newcommand{\eea}{\end{eqnarray}}
\newcommand{\beal}{\beq \begin{array}{ll}}
\newcommand{\eeal}{\end{array} \eeq}
\newcommand{\beas}{\begin{eqnarray*}}
\newcommand{\eeas}{\end{eqnarray*}}
\newcommand{\ba}{\begin{array}}
\newcommand{\ea}{\end{array}}
\newcommand{\bit}{\begin{itemize}}
\newcommand{\eit}{\end{itemize}}
\newcommand{\ben}{\begin{enumerate}}
\newcommand{\een}{\end{enumerate}}

\makeatletter
\DeclareRobustCommand\onedot{\futurelet\@let@token\@onedot}
\def\@onedot{.}
\def\eg{e.g\onedot} 
\def\ie{i.e\onedot} 
\def\cf{c.f\onedot}

\DeclareMathOperator*{\argmin}{argmin}
\DeclareMathOperator*{\argmax}{argmax}

\newcommand{\calXstar}{\calX^\star}

\begin{document}

\thispagestyle{empty}

\onecolumn
{\centering
This paper has been accepted for publication in IEEE Transactions on Robotics.
\vspace{5mm}

DOI: 10.1109/TRO.2016.2624754\\
IEEE Explore: \url{http://ieeexplore.ieee.org/document/7747236/}

\vspace{1cm}

Please cite the paper as:\\
\vspace{5mm}
C. Cadena and L. Carlone and H. Carrillo and Y. Latif and D. Scaramuzza and J. Neira and I. Reid and J.J. Leonard,\\
``Past, Present, and Future of Simultaneous Localization And Mapping: Towards the Robust-Perception Age'',\\
in IEEE Transactions on Robotics 32 (6) pp 1309-1332, 2016\\
}

\vspace{1cm}

bibtex:\\

\begin{footnotesize}
\begin{tabular}{llcl}
\multicolumn{4}{l}{@article\{Cadena16tro-SLAMfuture,}\\
&title &=& \{Past, Present, and Future of Simultaneous Localization And Mapping: Towards the Robust-Perception Age\},\\
&author &=& \{C. Cadena and L. Carlone and H. Carrillo and Y. Latif and D. Scaramuzza and J. Neira and I. Reid and J.J. Leonard\},\\
&journal &=& \{\{IEEE Transactions on Robotics\}\},\\
&year &=& \{2016\},\\
&number &=& \{6\},\\
&pages  &=& \{1309--1332\},\\
&volume &=& \{32\}\\
\}& & &
\end{tabular}
\end{footnotesize}

\twocolumn
\newpage
\setcounter{page}{1}

\title{Past, Present, and Future of Simultaneous Localization And Mapping: Towards the Robust-Perception Age}

\author{
Cesar~Cadena, Luca~Carlone, Henry~Carrillo, Yasir~Latif, \\ Davide~Scaramuzza, Jos\'e~Neira, Ian~Reid, John~J.~Leonard
\thanks{
C.\,Cadena is with the Autonomous Systems Lab, ETH Z\"{u}rich, Switzerland.
e-mail: \texttt{cesarc@ethz.ch}}%
\thanks{
L.\,Carlone is with the Laboratory for Information and Decision Systems, Massachusetts Institute of Technology, 
USA. 
e-mail: \texttt{lcarlone@mit.edu}}%
\thanks{
H.\,Carrillo is with the Escuela de Ciencias Exactas e Ingenier\'ia, Universidad Sergio Arboleda, 
Colombia, and Pontificia Universidad Javeriana, 
Colombia.
e-mail: \texttt{henry.carrillo@usa.edu.co}}%
\thanks{
Y.\,Latif and I.\,Reid are with the School of Computer Science, University of Adelaide, Australia, and the Australian Center for Robotic Vision.  
e-mail: \texttt{yasir.latif@adelaide.edu.au}, \texttt{ian.reid@adelaide.edu.au}}%
\thanks{
J.\,Neira is with the Departamento de Inform\'atica e Ingenier\'ia de Sistemas, Universidad de Zaragoza, 
Spain.
e-mail: \texttt{jneira@unizar.es}}%
\thanks{
D.\,Scaramuzza is with the Robotics and Perception Group, University of Z\"{u}rich, Switzerland.
e-mail: \texttt{sdavide@ifi.uzh.ch}}%
\thanks{
J.J.\,Leonard is with Marine Robotics Group, Massachusetts Institute of Technology, 
USA. 
e-mail: \texttt{jleonard@mit.edu}
}
\thanks{
This paper summarizes and extends the outcome of the workshop ``The Problem of Mobile Sensors: Setting future goals and indicators of progress for SLAM''~\cite{RSS15website}, held during the \emph{Robotics: Science and System} (RSS) conference (Rome, July 2015).
}
\thanks{
This work has been partially supported by the following grants: MINECO-FEDER DPI2015-68905-P, Grupo DGA T04-FSE; ARC grants DP130104413, CE140100016 and FL130100102; NCCR Robotics; PUJ 6601; EU-FP7-ICT-Project TRADR 609763, EU-H2020-688652 and SERI-15.0284.
}
}


\maketitle
\begin{abstract}
Simultaneous Localization And Mapping (SLAM)
 consists in the concurrent construction of a model of the environment (the \emph{map}), 
 and the estimation of the state of the robot moving within it.
The SLAM community has made astonishing progress over the last 30 years, enabling large-scale 
real-world applications, and witnessing a steady transition of this technology to industry. 
We survey the current state of SLAM and consider future directions. 
We start by presenting what is now the \emph{de-facto} standard formulation for SLAM.
We then review related work, covering a broad set of topics including robustness and scalability in long-term mapping, 
metric and semantic representations for mapping, theoretical performance guarantees, 
active SLAM and exploration, and other new frontiers. 
This paper simultaneously serves as a position paper and tutorial to
those who are users of SLAM.
By looking at the published research with a critical eye, 
we delineate open challenges and new research issues, that still deserve
careful scientific investigation.
The paper also contains the authors' take on two questions that often animate discussions 
during robotics conferences:
\emph{Do robots need SLAM?} and \emph{Is SLAM solved?}
\end{abstract}

\begin{IEEEkeywords}
Robots, SLAM, Localization, Mapping, Factor graphs, Maximum a posteriori estimation, 
sensing, perception.
\end{IEEEkeywords}
\blue{
\section*{Multimedia Material}
\noindent
Additional material for this paper, including an extended list of references (bibtex) and a table of 
pointers to   
online datasets for SLAM, can be found at \texttt{https://slam-future.github.io/}.
}

\newpage

\section{Introduction} 
\label{sec:intro}

\IEEEPARstart{S}{LAM} comprises the simultaneous estimation of the state of a robot equipped with on-board sensors,
 and the construction of a model (the \emph{map}) of the environment that the sensors are perceiving.
In simple instances, the robot state is described by its pose (position and orientation), although other quantities may be included in the state, such as 
robot velocity, sensor biases, and calibration parameters.
The map, on the other hand, is 
 a representation of aspects of interest (e.g., position of landmarks, obstacles)
 describing the environment in which the robot operates.

The need to \blue{use} a map of the environment  is twofold. First, the map is often required to support other tasks; for instance, 
a map can inform path planning or  provide an intuitive visualization for a human 
operator. Second, the map allows limiting the error committed in estimating the state of the robot. 
In the absence of a map, dead-reckoning would quickly drift over time; on the other hand, using a map, 
\blue{e.g., a set of distinguishable landmarks,}
the robot can ``reset'' its localization error by re-visiting known areas 
(so-called \emph{loop closure}).
Therefore, SLAM finds applications in all scenarios in which a prior map is not available and 
needs to be built. 

In some robotics applications \blue{the location of a set of landmarks} is known \textit{a priori}. For instance, a robot operating on a factory floor can be provided with a manually-built map of artificial beacons in the 
environment. Another example is the case in which the robot has access to GPS  
(the GPS satellites can be considered as moving beacons at known locations).
\blue{ 
In such scenarios, SLAM may not be required if localization can be done reliably with respect to the known landmarks.
}

The popularity of the SLAM problem is connected with the emergence of indoor applications of mobile robotics. 
Indoor operation rules out the use of GPS to bound the localization error; furthermore, SLAM 
provides an appealing alternative to user-built maps, showing that robot operation is 
possible in the absence of an 
ad hoc 
localization infrastructure.

A thorough historical review of the first 20 years of the SLAM problem is given by Durrant-Whyte and Bailey 
in two surveys~\cite{DurrantWhyte06ram,Bailey06ram}. 
These mainly cover what we call the 
 \emph{classical age} (1986-2004); 
 the classical age saw the introduction of the 
main probabilistic formulations for SLAM, including approaches based on 
Extended Kalman Filters, Rao-Blackwellised Particle Filters, and maximum likelihood estimation;
moreover, it delineated the basic challenges
connected to efficiency and robust data association. 
Two other excellent references describing the three main SLAM formulations of the classical age
are the book of Thrun, Burgard, and Fox~\cite{Thrun2005}
and \blue{the chapter of Stachniss\setal~\cite[Ch.~46]{Stachniss-HR2016}.}
The subsequent period is what we call the \emph{algorithmic-analysis age} (2004-2015), and 
is partially covered by 
Dissanayake\setal~in~\cite{dissanayake-review2011}. 
The algorithmic analysis period saw the study of fundamental properties of SLAM, 
including observability, convergence, and 
consistency. 
In this period, the key role of sparsity towards efficient SLAM solvers was also understood,
and the main open-source SLAM libraries were developed. 

We review the main SLAM surveys to date in~\prettyref{tab:surveys}, observing that most recent surveys only cover specific aspects or sub-fields of SLAM.
The popularity of SLAM in the last 30 years is not surprising if one thinks about the manifold aspects that  SLAM involves.  
At the lower level (called the \emph{front-end} in Section~\ref{sec:problemFormulation}) SLAM naturally intersects other research fields such as computer vision and signal processing; at the higher level (that we later call the \emph{back-end}), SLAM is an appealing mix of geometry, graph theory, optimization, and probabilistic estimation.
Finally, a SLAM expert has to deal with practical aspects ranging from sensor \blue{calibration} to system integration.

The present paper gives a broad overview of the current state of SLAM, and offers the perspective of part of the community on the open problems and future directions for the SLAM research.
Our main focus is on metric and semantic SLAM, and we refer the reader to the recent survey by Lowry\setal~\cite{lowry-tro2016rohtua}, which provides a comprehensive review of vision-based place recognition and topological SLAM.

Before delving into the paper, we first discuss two questions that
often animate discussions during robotics conferences: \blue{ (1) do
  autonomous robots need SLAM? and (2) is SLAM solved as an academic
  research endeavor?  We will revisit these questions at the end of
the manuscript.}


%

\notAddressed{Guys, either the question is wrong or the answer is wrong: The text in this section does not explaun why robots need SLAM. 
the reasons should be explained from the perspective of the requirements of robots, not from the perspective of research, unless you change the question.}

Answering the question ``Do autonomous robots really need SLAM?''
requires understanding what makes SLAM unique.  SLAM aims at building
a globally consistent representation of the environment, leveraging
both ego-motion measurements and loop closures.  The keyword here is
``loop closure'': if we sacrifice loop closures, SLAM reduces to
odometry.  In early applications, odometry was obtained by integrating
wheel encoders.  The pose estimate obtained from wheel odometry
quickly drifts, making the estimate unusable after few
meters~\cite[Ch. 6]{Kelly2013}; this was one of the main thrusts
behind the development of SLAM: the observation of external landmarks
is useful to reduce the trajectory drift and possibly correct
it~\cite{Newman2002a}. However, more recent odometry algorithms are
based on visual and inertial information, and have very small drift
($<0.5\%$ of the trajectory length~\cite{Forster15arxiv}).  Hence the
question becomes legitimate: do we really need SLAM?  Our answer is
three-fold.

First of all, we observe that the SLAM research 
done over the last decade has itself produced the visual-inertial odometry algorithms 
that currently represent the state of the art, e.g.,~\cite{Mourikis07icra,Lynen-RSS-15}; 
in this sense Visual-Inertial Navigation (VIN) \emph{is} SLAM: 
VIN can be  considered a \emph{reduced} SLAM system, in which the loop closure 
(or place recognition) module is disabled. More generally, SLAM has directly led to the study of  
sensor fusion under more challenging setups (i.e., no GPS, low quality sensors) 
than previously considered in other literature 
(e.g., inertial navigation in aerospace engineering\notAddressed{Citation to support this claim?}).


\setlength\extrarowheight{5pt}

\begin{table}[t]
  \centering
  \caption{Surveying the surveys and tutorials \label{tab:surveys}}
  \label{tab:survey}
 \begin{tabular}{| m{0.4cm} | m{3cm} | m{4.3cm} |}
 \hline
 {\bf Year} & {\bf Topic} & {\bf Reference} \\
 \hline\hline
 2006
 & Probabilistic approaches and data association
 & Durrant-Whyte and Bailey~\cite{DurrantWhyte06ram,Bailey06ram}
 \\
 \hline
 2008
 & Filtering approaches 
 & Aulinas\setal~\cite{Aulinas2008survey}
 \\
 \hline
 2011
 & SLAM back-end
 & Grisetti\setal~\cite{grisetti2010tutorial}
 \\
 \hline
 2011
 & Observability, consistency and convergence
 & Dissanayake\setal~\cite{dissanayake-review2011}
 \\
  \hline
 2012
 & Visual odometry
 & Scaramuzza and Fraundofer~\cite{Scaramuzza11ram,Fraundorfer12ram}
 \\
   \hline
 2016
 & Multi robot SLAM
 & Saeedi\setal~\cite{saeedi-jfr2016}
 \\
  \hline
 2016
 & Visual place recognition
 & Lowry\setal~\cite{lowry-tro2016rohtua}
 \\
  \hline
   2016
 & SLAM in the Handbook of Robotics
 & Stachniss\setal~\cite[Ch. 46]{Stachniss-HR2016}
 \\
  \hline
   2016
 & Theoretical aspects
 & Huang and Dissanayake~\cite{huang2016ijars}
 \\
  \hline  
\end{tabular}
\vspace{-5mm}
\end{table}

\begin{figure}
\centering
\includegraphics[width=\columnwidth]{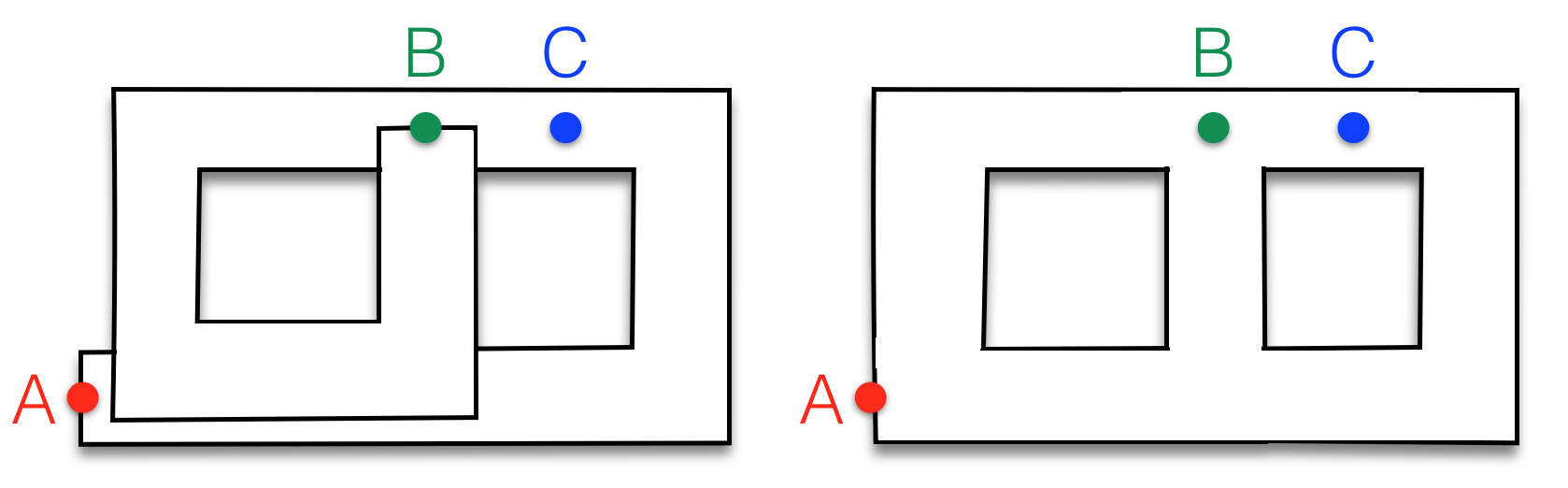}
\caption{Left: map built from odometry. The map is homotopic to a long corridor 
that goes from the starting position A to the final position B. Points that are close 
in reality (e.g., B and C) may be arbitrarily far in the odometric map.
Right: map build from SLAM. By leveraging loop closures, 
SLAM estimates the actual topology of the environment, 
and ``discovers'' shortcuts in the map.}
\label{fig:topology}
\vspace{-5mm}
\end{figure}

\YL{The second answer regards the true topology of the environment.}
A robot performing odometry and neglecting loop closures interprets the world as an 
``infinite corridor'' (\prettyref{fig:topology}-left) 
in which the robot keeps exploring new areas indefinitely. 
A loop closure event informs the robot that this ``corridor'' 
keeps intersecting itself (\prettyref{fig:topology}-right). The advantage of loop closure now becomes clear:
by finding loop closures, the robot understands the real topology of the environment, 
and is able to find shortcuts between locations (e.g., point B and C in the map).
Therefore, if getting the right topology of the environment is one of the merits of SLAM, 
why not simply drop the metric information and just do 
place recognition?
The answer is simple: the metric information makes place recognition much simpler and more robust; 
the metric reconstruction informs the robot about loop closure opportunities and allows 
discarding spurious loop closures~\cite{Latif2013ijrr}. 
Therefore, while SLAM might be redundant in principle
 (an oracle place recognition module would suffice for topological mapping), 
 SLAM offers a natural defense against wrong 
 data association and perceptual aliasing, where similarly looking scenes, corresponding to 
 distinct locations in the environment, would deceive place recognition. 
 In this sense, the SLAM map provides a way to predict and validate future measurements:
we believe that this mechanism is  key to robust operation.

The third answer is that SLAM is needed for many applications that,
either implicitly or explicitly, \emph{do} require a globally
consistent map.  For instance, in many military and civilian
applications, the goal of the robot is to explore an environment and
report a map to the human operator, \blue{ensuring that full coverage of the
environment has been obtained.}  Another example is the case in which
the robot has to perform structural inspection (of a building, bridge,
etc.); also in this case a globally consistent 3D reconstruction is a
requirement for successful operation.



This question of ``is SLAM solved?'' is often asked within the robotics community, c.f.~\cite{Frese2010a}.
This question is difficult to answer because SLAM has become such a broad topic that the question is well posed only for a  given robot/environment/performance combination. 
In particular, one can evaluate the maturity of the SLAM problem once the following aspects are 
specified:
\bit
\item \emph{robot}: type of motion (e.g., dynamics, maximum speed),
 available sensors (e.g., resolution, sampling rate), 
 available computational resources;
\item \emph{environment}: planar or three-dimensional, 
presence of natural or artificial landmarks,
amount of dynamic elements,  
amount of symmetry and risk of perceptual aliasing.
Note that many of these aspects 
actually depend on the sensor-environment pair: for instance,
two rooms may look identical for a 2D laser scanner (perceptual aliasing), 
while a camera may discern them from appearance cues; 
\item \emph{performance requirements}: desired accuracy in the estimation of the state of the robot, 
accuracy and type of representation of the environment (e.g., landmark-based or dense), 
success rate (percentage of tests in which the accuracy bounds are met), 
estimation latency, maximum operation time, maximum size of the mapped area.
\eit
For instance, mapping a 2D indoor environment with a robot equipped with wheel encoders and a 
laser scanner, with sufficient accuracy ($<10$cm) and sufficient robustness (say, low failure rate), 
can be considered largely solved (an example of industrial system performing SLAM is the \emph{Kuka Navigation 
Solution}~\cite{kukaSLAM}).
Similarly, vision-based SLAM with slowly-moving robots (e.g., Mars rovers~\cite{Maimone-JFR07}, 
domestic robots~\cite{dysonSLAM}), 
and visual-inertial odometry~\cite{tangoGoogle} can be considered mature research fields. 


On the other hand, other robot/environment/performance combinations
still deserve a large amount of fundamental research.  Current SLAM
algorithms can be easily induced to fail when either the motion of the
robot or the environment are too challenging (e.g., fast robot
dynamics, highly dynamic environments); similarly, SLAM algorithms are
often unable to face strict performance requirements, e.g., high rate
estimation for fast closed-loop control.  This survey will provide a
comprehensive overview of these open problems, among others.

In this paper, we argue that we are entering in a third era for SLAM, the  
\emph{robust-perception age}, which is characterized by the following key requirements:
\ben
\item \emph{robust performance}: the SLAM system operates with low failure rate 
for an extended period of time in a broad set of environments; 
the system includes fail-safe mechanisms and has 
self-tuning capabilities\footnote{The SLAM community 
has been largely affected by the ``curse of manual tuning'', in that satisfactory operation 
is enabled by expert tuning of the system parameters 
(e.g., stopping conditions, thresholds for outlier rejection).} 
in that it can adapt the selection 
of the system parameters to the scenario.
\item \emph{high-level understanding}: the SLAM system goes beyond basic geometry reconstruction 
to obtain a high-level understanding of the environment (e.g., high-level geometry, semantics, physics, affordances);
\item \emph{resource awareness}: the SLAM system is tailored to the available sensing and 
computational resources, and provides means to adjust the computation load depending on the 
available resources;
\item \emph{task-driven perception}: the SLAM system is able to select relevant 
perceptual information and filter out irrelevant sensor data, in order to support 
the task the robot has to perform; moreover,
the SLAM system produces adaptive map representations, whose 
complexity may vary depending on the task at hand. 
\een

\myParagraph{Paper organization}
The paper starts by presenting a standard formulation and architecture for SLAM
(Section~\ref{sec:problemFormulation}).
Section~\ref{sec:robustness} tackles robustness in life-long SLAM.
Section~\ref{sec:scalability} deals with scalability.
Section~\ref{sec:rep-metric} discusses how
to represent the geometry of the environment. Section~\ref{sec:rep-semantic} extends the 
question of the environment representation to the modeling of semantic information.
Section~\ref{sec:theory} provides an overview of the current accomplishments on the 
theoretical aspects of SLAM. 
Section~\ref{sec:ActiveSLAM} broadens the discussion and reviews the active SLAM problem in which 
decision making is used to improve the quality of the SLAM results.
Section~\ref{sec:newTrends} provides an overview of recent trends in SLAM, including 
the use of unconventional sensors \blue{and deep learning}. 
Section~\ref{sec:conclusion} provides final remarks. 
Throughout the paper, we provide many pointers to related work outside the robotics community.
Despite its unique traits, SLAM is related to problems addressed in computer vision, computer 
graphics, and control theory, and cross-fertilization among these fields is a necessary condition
to enable fast progress. 


For the non-expert reader, we recommend to read Durrant-Whyte and Bailey's SLAM tutorials \cite{DurrantWhyte06ram,Bailey06ram} before delving in this 
position paper. The more experienced researchers can jump directly to the section of interest, where they will find a self-contained overview of 
the state of the art and open problems. 

\newcommand{\Prob}{\text{p}} 
\newcommand{\hfun}{h}
\newcommand{\meas}{z}
\newcommand{\Cov}{\Sigma}
\newcommand{\Info}{\Omega}
\newcommand{\calXhat}{\hat{\calX}}
\newcommand{\deltaX}{\delta_{\calX}}
\newcommand{\deltaXstar}{\delta^\star_{\calX}}
\newcommand{\jac}{A}
\newcommand{\rhs}{b}
\newcommand{\GN}{GN\xspace}
\newcommand{\nrMeas}{m}


\begin{figure}
\includegraphics[width=1.02\columnwidth]{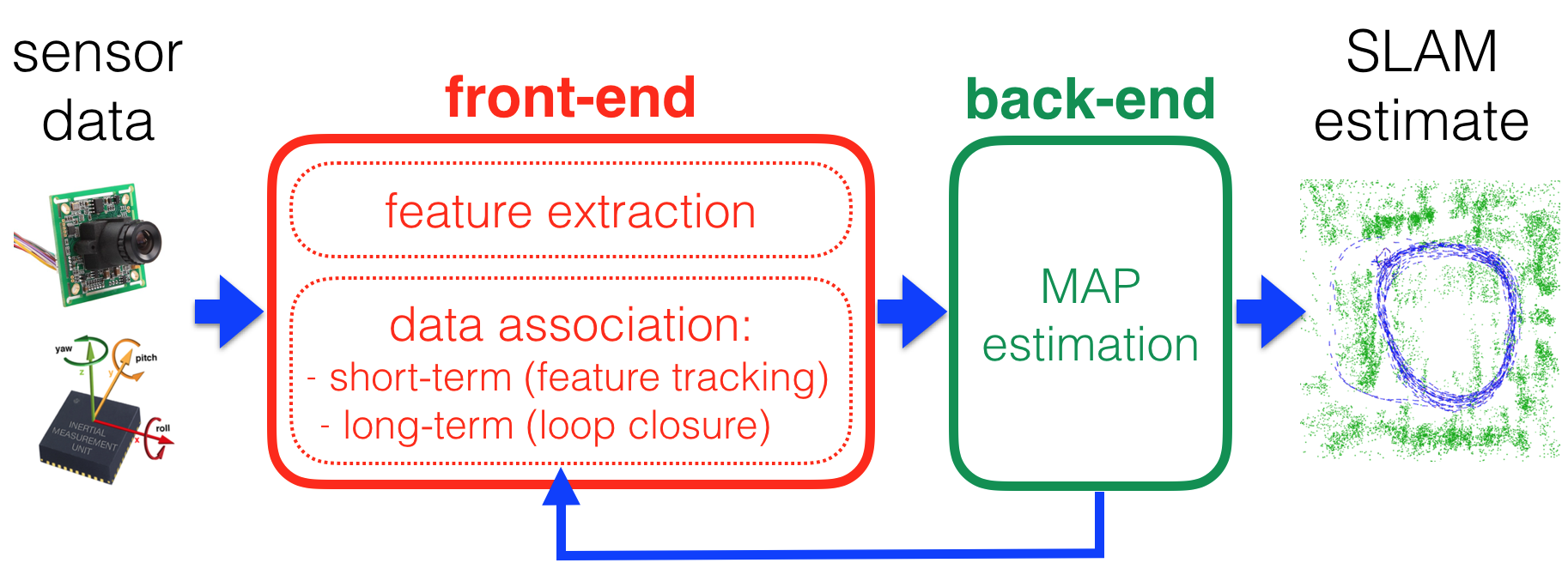} 
\caption{
Front-end and back-end in a typical SLAM system. \YL{The back-end can provide feedback to the front-end for loop closure detection and verification.}
\label{fig:frontBack}
}
\end{figure}

\section{Anatomy of a Modern SLAM System}
\label{sec:problemFormulation}

The architecture of a SLAM system includes two main components: the \textit{front-end} and the \textit{back-end}.
The front-end abstracts sensor data into models that are amenable for estimation, while the 
back-end performs inference on the abstracted data produced by the front-end. This architecture is summarized in~\prettyref{fig:frontBack}.
We review both components, starting from the back-end. 

\myParagraph{Maximum a posteriori (MAP) estimation and the SLAM back-end}
The current \textit{de-facto} standard formulation of  SLAM  has its origins in the seminal paper of Lu and Milios~\cite{Lu-AutRob97}, followed by the work of Gutmann and Konolige \cite{Gutmann-99}.
Since then, numerous approaches have improved the efficiency and robustness of the optimization  
underlying the problem \cite{Thrun05ijrr-graphSLAM,
Folkesson-JFR06, Dellaert-IJRR06, Olson06icra, Grisetti-RSS07,Kaess-ijrr2012}. 
All these approaches
formulate SLAM  
as a maximum a posteriori estimation problem,  
and often use the formalism of \textit{factor graphs}~\cite{Kschischang01it} to reason about 
the interdependence among variables.

Assume that we want to estimate an unknown variable $\calX$; as mentioned before, 
in SLAM the variable $\calX$ typically includes the 
trajectory of the robot (as a discrete set of poses) and the position of landmarks in the environment.
We are given a set of measurements $Z = \{z_k: k=1,\ldots,\nrMeas\}$ 
such that each measurement can be expressed as a function of $\calX$, i.e., 
$\meas_k = \hfun_k(\calX_k) + \epsilon_k$, where $\calX_k \subseteq \calX$ is a 
subset of the variables, $\hfun_k(\cdot)$ is a known function (the \emph{measurement} or 
\emph{observation} model), 
and $\epsilon_k$ is random measurement noise. 

In MAP estimation, we estimate $\calX$ by computing the 
 assignment of variables $\calXstar$ that attains the maximum of the posterior  
 $\Prob(\calX|Z)$ (the \emph{belief} over $\calX$ given the measurements):
\begin{equation}
\calXstar \doteq \;  
\; \argmax_{\calX} \; \Prob(\calX|Z) =
 \argmax_{\calX} \; \Prob(Z|\calX) \Prob(\calX)
\label{equ:generalFGMax0}
\end{equation}
where the equality follows from the Bayes theorem. 
In~\eqref{equ:generalFGMax0}, $\Prob(Z|\calX)$ is the likelihood of the measurements $Z$ given the 
assignment $\calX$, and $\Prob(\calX)$ is a prior probability over $\calX$. 
The prior probability
includes any prior knowledge about $\calX$; in case no prior knowledge is available, 
$\Prob(\calX)$ becomes a constant (uniform distribution) which is inconsequential and can be dropped 
from the optimization. In that case MAP estimation reduces to \emph{maximum likelihood estimation}.
\blue{Note that, unlike Kalman filtering, MAP estimation does not require 
an explicit distinction between motion and observation model: 
both models are treated as factors and are seamlessly incorporated in the estimation process.
Moreover, it is worth noting that Kalman filtering and MAP estimation return the 
same estimate in the linear Gaussian case, while this is not the case in general.}

Assuming that the measurements $Z$
are independent (i.e., the 
corresponding noises are uncorrelated), problem~\eqref{equ:generalFGMax0}
 factorizes into:
\bea
\calXstar = \argmax_{\calX} \; \Prob(\calX) \prod_{k=1}^\nrMeas \Prob(\meas_k|\calX) = \nonumber \\ 
\argmax_{\calX} \; \Prob(\calX) \prod_{k=1}^\nrMeas \Prob(\meas_k|\calX_k) 
\label{equ:generalFGMax}
\eea
where, on the right-hand-side, we noticed that $\meas_k$ only depends on the subset of variables
in $\calX_k$.

\begin{figure}
\centering
\includegraphics[width=0.7\columnwidth]{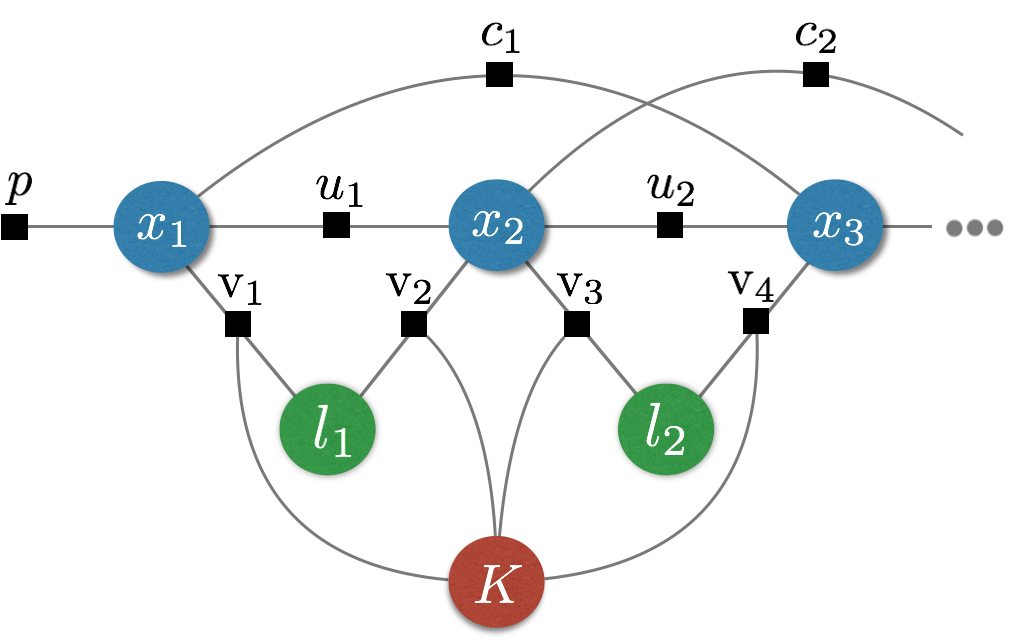} 
\caption{\textbf{SLAM as a factor graph}:
Blue circles denote robot poses at consecutive time steps ($x_1, x_2, \ldots$), 
green circles denote landmark positions ($l_1, l_2, \ldots$), 
red circle denotes the variable associated with the 
intrinsic calibration parameters ($K$). Factors are shown as black squares: 
the label ``u'' marks factors corresponding to odometry constraints, 
``v'' marks factors corresponding to camera observations,
``c'' denotes loop closures, and 
``p'' denotes prior factors.
\label{fig:factorGraph}
}
\vspace{-5mm}
\end{figure}

Problem~\eqref{equ:generalFGMax} can be interpreted in terms of inference over a 
factors graph~\cite{Kschischang01it}. The variables correspond to nodes in the factor graph.
The terms $\Prob(\meas_k|\calX_k)$ and the prior 
$\Prob(\calX)$ are called \emph{factors}, and they encode probabilistic constraints
over a subset of nodes.
 A factor graph is a graphical model
 that encodes the dependence between the $k$-th factor (and its measurement $\meas_k$) and the corresponding variables $\calX_k$. 
A first advantage of the factor graph interpretation is that it enables an insightful visualization of the 
problem. \prettyref{fig:factorGraph} shows an example of a factor graph underlying a simple SLAM problem. 
The figure shows the variables, namely, the robot poses, the landmark positions, 
and the camera calibration parameters, and the factors imposing constraints among these variables.
A second advantage is generality: a factor graph can model complex inference problems with heterogeneous variables and factors, and arbitrary interconnections. \blue{Furthermore,} the connectivity of the factor graph 
in turn influences the sparsity of the resulting SLAM problem as discussed below.
%
%

In order to write~\eqref{equ:generalFGMax} in a more explicit form,
 assume that the measurement noise
 $\epsilon_k$ is a zero-mean Gaussian noise with information matrix $\Info_k$ (inverse of the covariance matrix).
 Then, the measurement likelihood in~\eqref{equ:generalFGMax} becomes:
\begin{equation}
\label{eq:measLike}
\Prob(\meas_k|\calX_k) \propto \text{exp}(-\frac{1}{2} || \hfun_k(\calX_k) - \meas_k ||^2_{\Info_k})
\end{equation}
\noindent
where 
we use the  notation
$||e||^2_{\Info} = e\tran\Info e$. 
Similarly, assume that the prior can be written as: 
%
$\Prob(\calX) \propto \text{exp}(-\frac{1}{2} || \hfun_0(\calX) - \meas_0 ||^2_{\Info_0})$,
for some given function $\hfun_0(\cdot)$, prior mean $\meas_0$, and information matrix $\Info_0$.
Since maximizing the posterior is the same as minimizing the \emph{negative log-posterior}, the MAP estimate 
in~\eqref{equ:generalFGMax} becomes:
\bea
\calXstar = 
\argmin_{\calX} -\text{log} \left( \Prob(\calX) \prod_{k=1}^\nrMeas \Prob(\meas_k|\calX_k)  \right) = \nonumber \\
\argmin_{\calX} \sum_{k=0}^\nrMeas || \hfun_k(\calX_k) - \meas_k ||^2_{\Info_k}
\label{NonLinearLS}
\eea
\noindent which is a nonlinear least squares problem, as in most problems of interest in robotics, 
$\hfun_k(\cdot)$ is a nonlinear function.
Note that the formulation~\eqref{NonLinearLS} follows from the assumption of Normally 
distributed noise. Other assumptions for the noise distribution lead to 
different cost functions; for instance, if the noise follows a Laplace distribution,
the squared $\ell_2$-norm in~\eqref{NonLinearLS} is replaced by the $\ell_1$-norm.
To increase resilience to outliers, it is also common to substitute 
the squared $\ell_2$-norm in~\eqref{NonLinearLS} with  
robust loss functions (e.g., Huber or Tukey loss)~\cite{huber1981robust}.


The computer vision expert may notice a resemblance between problem~\eqref{NonLinearLS}
 and \emph{bundle adjustment} (BA) 
in Structure from Motion~\cite{Triggs00}; both~\eqref{NonLinearLS} and BA indeed stem from a 
maximum a posteriori formulation. However, two key features make SLAM unique. 
First, the factors in~\eqref{NonLinearLS} are not constrained to model projective geometry as in BA, 
but include a broad variety of sensor models, e.g., inertial sensors, wheel encoders, 
GPS, to mention a few. 
For instance, in laser-based mapping, the factors usually constrain  relative 
poses corresponding to different viewpoints, while in direct
 methods for visual SLAM, the factors penalize differences in pixel intensities 
across different views of the same portion of the scene.
The second difference with respect to BA is that, in a SLAM scenario, 
 problem~\eqref{NonLinearLS} typically needs to be solved \emph{incrementally}: 
new measurements are made available at each time step as the robot moves.

\blue{The minimization problem~\eqref{NonLinearLS} is commonly solved via successive linearizations, 
e.g., the Gauss-Newton or the Levenberg-Marquardt methods (alternative approaches, 
based on convex relaxations and Lagrangian duality are reviewed in Section~\ref{sec:theory}). 
Successive linearization methods proceed iteratively, starting from a given initial guess $\calXhat$, 
and approximate the cost function at $\calXhat$ with a quadratic cost, which can be optimized in 
closed form by solving a set of linear equations (the so called \emph{normal equations}). 
 These approaches can be seamlessly generalized to variables 
 belonging to smooth manifolds (e.g., rotations), which are of interest in robotics~\cite{Absil07book,Forster15arxiv}.

The key insight behind modern SLAM solvers is that the matrix appearing in the normal equations 
 is sparse and  its sparsity structure is dictated by the topology of the underlying factor graph.
This enables the use of fast linear solvers~\cite{Kaess08tro,Kuemmerle11icra,Kaess-ijrr2012,Polok13rss}.
Moreover, it allows designing \emph{incremental} (or \emph{online}) solvers, which update the estimate of $\calX$ 
 as new observations are acquired~\cite{Kaess08tro,Kaess-ijrr2012,Polok13rss}. 
 Current SLAM libraries 
 (e.g., GTSAM\,\cite{Dellaert12tr}, g2o\,\cite{Kuemmerle11icra}, Ceres\,\cite{ceres}, iSAM\,\cite{Kaess08tro}, 
 and SLAM++\,\cite{Polok13rss}) are able to solve problems with tens of thousands of variables in few seconds. 
The hands-on tutorials~\cite{Dellaert12tr,grisetti2010tutorial} provide excellent
  introductions to two of the 
 most popular SLAM libraries; each library also includes a set of examples showcasing  
 real SLAM problems.

The SLAM formulation described so far is commonly referred to as 
\emph{maximum a posteriori} 
estimation,
\emph{factor graph optimization}, \emph{graph-SLAM}, 
 \emph{full smoothing}, or \emph{smoothing and mapping} (SAM). 
A popular variation of this framework is \emph{pose graph optimization},
in which
the variables to be estimated are 
poses sampled along the trajectory of the robot, and each factor imposes a constraint on a pair of poses.}

MAP estimation has been proven to be more accurate and efficient 
than original approaches for SLAM based on nonlinear filtering. 
We refer the reader to the surveys~\cite{DurrantWhyte06ram,Bailey06ram} for an overview 
on filtering approaches, and to~\cite{Strasdat-ivc2012}
for a comparison between filtering and smoothing. 
We remark that some SLAM systems based on EKF have also been demonstrated to attain 
state-of-the-art performance. Excellent examples of EKF-based SLAM systems include the 
Multi-State Constraint {K}alman Filter of Mourikis and Roumeliotis~\cite{Mourikis07icra}, 
and the VIN systems of Kottas\setal~\cite{Kottas2012iser} and Hesch\setal~\cite{Hesch14ijrr}. 
Not surprisingly, the performance mismatch between filtering and MAP estimation 
gets smaller when the linearization point for the EKF is accurate 
(as it happens in visual-inertial navigation problems), 
when using sliding-window filters, and when potential sources of 
inconsistency in the EKF are taken care of~\cite{Huang11iros,Hesch14ijrr,Kottas2012iser}.   

As discussed in the next section, MAP estimation is usually performed on 
a pre-processed version of the sensor data. 
In this regard, it is often referred to as the SLAM \emph{back-end}.


\myParagraph{Sensor-dependent SLAM front-end}
In practical robotics applications, it might be hard to write directly the sensor 
measurements as an analytic function of the state, as required in MAP estimation. 
For instance, if the raw sensor data is an image, it might be hard to express 
the intensity of each pixel as a function of the SLAM state; the same difficulty arises 
 with simpler sensors (e.g., a laser with a single beam). In both cases the issue is 
connected with the fact that we are not able to design a sufficiently general, yet 
tractable representation of the environment; even in the presence of such a general representation, 
it would be hard to write an analytic function that connects the measurements to the 
parameters of such a representation.

For this reason, before the SLAM back-end, 
it is common to have a module, the  \emph{front-end}, that extracts relevant features from the sensor data.
 For instance, in vision-based SLAM, the front-end 
extracts the pixel location of few distinguishable points in the environment; pixel 
observations of these points are now easy to model within the back-end. 
The front-end is also in charge of associating each measurement to a specific 
landmark (say, 3D point) in the environment: this is the so called \emph{data association}.
More abstractly, the data association 
module associates each measurement $\meas_k$ with a subset of unknown variables $\calX_k$ such
that $\meas_k = h_k(\calX_k)+\epsilon_k$.
Finally, the front-end might also provide an initial guess for the variables in the nonlinear 
optimization~\eqref{NonLinearLS}. For instance, in feature-based monocular SLAM 
the front-end usually takes care of the landmark initialization, by triangulating 
the position of the landmark from multiple views.

A pictorial representation of a typical SLAM system is given in~\prettyref{fig:frontBack}.
The data association module in the front-end includes a short-term data association block 
and a long-term one. Short-term data association is responsible for associating 
corresponding features in consecutive sensor measurements; for instance,  
short-term data association would track the fact that 2 pixel measurements in consecutive 
frames are picturing the same 3D point. 
On the other hand, long-term data association (or loop closure) is in charge of associating 
new measurements to older landmarks. We remark that the back-end usually feeds back 
information to the front-end, e.g., to support loop closure detection and validation.

The pre-processing that happens in the front-end is sensor dependent, since the notion 
of \emph{feature} changes depending on the input data stream we consider. 


\section{Long-term autonomy I: Robustness} 
\label{sec:robustness}

A SLAM system might be fragile in many aspects: failure can be algorithmic\footnote{We 
omit the (large) class of software-related failures. The non-expert reader 
must be aware that integration and testing are key aspects of SLAM and robotics in general.
} or
hardware-related. The former class includes failure modes induced by limitation of the existing SLAM 
algorithms (e.g., difficulty to handle extremely dynamic or harsh environments). The latter includes 
failures due to sensor or actuator degradation. 
Explicitly addressing these failure modes is crucial 
for long-term operation, where one can no longer make \blue{simplifying} assumptions about the structure of 
the environment (e.g., mostly static) or fully rely on on-board sensors. 
In this section we review the main challenges to algorithmic robustness. 
We then discuss open problems, including 
robustness against  hardware-related failures.

One of the main sources of algorithmic failures is data association.  
As mentioned in Section~\ref{sec:problemFormulation} data association matches each measurement to 
the portion of the state the measurement refers to.
For instance, in feature-based visual SLAM, it associates each visual feature to a specific landmark. 
Perceptual aliasing, the phenomenon \YL{in which} different sensory inputs lead to the same sensor signature, 
makes this problem particularly hard. In the presence of perceptual aliasing, data association establishes erroneous 
measurement-state matches (outliers, or false positives), which 
\YL{in turn result} in wrong 
estimates from the back-end.
On the other hand, when data association decides to incorrectly reject 
a sensor measurement as spurious (false negatives), fewer measurements are used for 
estimation, at the expense of estimation accuracy.

The situation is made worse  by the presence of unmodeled dynamics in the environment including both short-term and seasonal changes, 
which might deceive short-term and long-term data association.
A  fairly common assumption in current SLAM approaches is that  
the world remains unchanged as the robot moves through it (in other words, landmarks are static). 
This  \textit{static world assumption} 
holds true in a single mapping run in small scale scenarios, 
as long as there are no \textit{short term dynamics} (e.g., people and objects moving around). 
 When mapping over longer time scales and in large environments, change is inevitable. 

Another aspect of robustness is that of doing SLAM in harsh environments such 
as underwater \cite{Eustice06ijrr,kim2013real}. 
The challenges in this case are the limited visibility, the constantly changing conditions, 
and the impossibility of using conventional sensors (e.g., laser range finder).

\myParagraph{Brief Survey}
Robustness issues connected to incorrect data association can be addressed in the front-end and/or in the back-end 
of a SLAM system.
Traditionally, the front-end has been entrusted with establishing correct data association.
Short-term data association is the easier one to tackle: if the sampling rate of the sensor is relatively fast, 
compared to the dynamics of the robot, tracking features that correspond to the same 3D landmark is easy.
For instance, if we want to track 
a 3D point across consecutive images and assuming that the framerate is sufficiently high, standard approaches based on 
 descriptor matching or optical flow~\cite{Scaramuzza11ram} ensure reliable tracking.
  Intuitively, at high framerate, the viewpoint of the sensor (camera, laser) does not change significantly, 
 hence the features at time $t+1$ \blue{(and its appearance)} remain close to the ones observed at time $t$.\footnote{In hindsight, the 
 fact that short-term data association is much easier and more reliable than the long-term one is what makes (visual, inertial) odometry 
 simpler than SLAM.}
Long-term data association in the front-end is more challenging and involves loop closure \emph{detection} and \emph{validation}. 
For loop closure detection at the front-end, 
a brute-force approach which detects features in the current measurement (e.g., image) and tries to match 
them against all previously detected features quickly becomes impractical.
Bag-of-words models \cite{Sivic03iccv} avoid this intractability by quantizing the feature space and allowing more efficient search.
Bag-of-words can be arranged into hierarchical vocabulary trees \cite{Nister06cvpr} that 
enable efficient lookup in large-scale datasets. Bag-of-words-based techniques such as 
\cite{Cummins08ijrr} have shown reliable performance on the task of single session loop closure detection.
 However, these approaches are not capable of handling severe illumination variations as visual words can no longer be matched. 
This has led to develop new methods that explicitly account for such variations by matching sequences \cite{Milford12icra}, gathering different visual appearances into a unified representation \cite{churchill2013}, or using spatial as well as appearance information \cite{ho2006loop}.
A detailed survey on visual place recognition can be found in Lowry\setal~\cite{lowry-tro2016rohtua}.
Feature-based methods have also been used to detect loop closures in laser-based SLAM front-ends; for instance, 
Tipaldi\setal~\cite{tipaldi2010flirt} propose FLIRT features for 2D laser scans.

Loop closure validation, instead, consists of additional geometric verification steps to ascertain the quality of the loop closure.
In vision-based applications, RANSAC is commonly used for geometric verification and outlier rejection, see~\cite{Scaramuzza11ram} 
and the references therein. In laser-based approaches, one can validate a loop closure by 
checking how well the current laser scan matches the existing map (i.e., how small is the residual error resulting from scan matching).

Despite the progress made to robustify loop closure detection at the front-end, 
in presence of perceptual aliasing, it is unavoidable that wrong loop closures 
are fed to the back-end.
Wrong loop closures can severely corrupt 
the quality of the MAP estimate~\cite{Sunderhauf12icra}.
In order to deal with this issue, 
a recent line of research~\cite{Latif2013ijrr,Sunderhauf12icra,Olson2013ijrr,Carlone14iros} 
proposes techniques to make the SLAM back-end resilient against spurious measurements. 
These methods reason on the validity of loop closure constraints by looking at the 
residual error induced by the constraints 
during optimization. Other methods, instead, attempt to detect outliers \textit{a priori}, that is, before any optimization takes place, by identifying incorrect loop closures that are not supported by the odometry \cite{sabatta2010improved}.



In dynamic environments, the challenge is twofold. 
First, the SLAM system has to detect, discard, or track
changes. While mainstream approaches attempt to discard the dynamic portion of the scene
\cite{Neira01tra}, some works include dynamic elements 
as part of the model~\cite{Bibby-RSS-07,Wang07ijrr}.
The second challenge regards the fact that the SLAM system has to model permanent or semi-permanent changes, 
and understand how and when to update the map accordingly.
Current SLAM systems that deal with dynamics either maintain 
multiple (time-dependent) maps of the same location
\cite{Dayoub2011a}, 
or have a 
single representation parameterized by some time-varying parameter 
\cite{krajnik-iros2014}. 



\myParagraph{Open Problems}
In this section we review open problems and  novel research questions arising in long-term SLAM.

\mySubParagraph{Failsafe SLAM and recovery} 
 Despite the progress made on the SLAM back-end,
 current SLAM solvers are still vulnerable in the presence of outliers. 
 This is mainly due to the fact that virtually all robust SLAM techniques are based on iterative 
 optimization of nonconvex costs. This has two consequences: first, the outlier rejection outcome depends on the quality 
 of the initial guess fed to the optimization; second, the system is inherently fragile:
 the inclusion of a single outlier  
 degrades the quality of the estimate, which in turn degrades the capability of 
 discerning outliers later on. 
 \YL{These types of failures lead to an incorrect linearization point from which recovery is not trivial, especially in an incremental setup.}
An ideal SLAM solution should be \textit{fail-safe} and \textit{failure-aware}, i.e.,
the system needs to be aware of imminent failure (e.g., due to outliers or degeneracies)
and provide recovery mechanisms that can re-establish proper operation. 
None of the existing SLAM approaches provides these capabilities.
\YL{A possible way to achieve this is a tighter integration between the front-end and the back-end, but how to achieve that is still an open question.}

\mySubParagraph{Robustness to HW failure} While addressing hardware failures might appear outside the scope
of SLAM, these failures impact the SLAM system, and the latter can play a key role in detecting and mitigating sensor and locomotion failures.
If the accuracy of a sensor degrades due to malfunctioning, \blue{off-nominal conditions}, or aging, the quality of the sensor measurements 
(e.g., noise, bias) does not match the noise model used in the back-end (\cf~eq.~\eqref{eq:measLike}), leading to poor estimates.
This naturally poses different research questions: how can we detect degraded sensor operation?
how can we adjust sensor noise statistics (covariances, biases) accordingly?
\blue{more generally, how do we resolve conflicting information from different sensors? 
This seems crucial in safety-critical applications (e.g., self-driving cars) in which misinterpretation 
of sensor data may put human life at risk.
}

\mySubParagraph{Metric Relocalization}
While appearance-based, as opposed to feature-based, methods are able to close loops between day and night sequences 
or between different seasons, 
the resulting loop closure is topological in nature.
For metric relocalization (i.e., estimating the relative pose with respect to the previously built map), feature-based approaches are still the norm; 
\blue{however, current feature descriptors lack sufficient invariance to work reliably under such circumstances.
Spatial information, inherent to the SLAM problem, such as trajectory matching}, might be exploited to overcome these limitations.
Additionally, mapping with one sensor modality (e.g., 3D lidar) 
and localizing in the same map with a different sensor modality  
(e.g., camera) can be a useful addition. 
 The work of Wolcott\setal~\cite{rwolcott-2014a} 
 is an \YL{initial} step in this direction.

\mySubParagraph{Time varying and deformable maps} Mainstream SLAM methods have been developed with the rigid and static world assumption in mind; however, the real world is non-rigid both due to dynamics as well 
as the inherent deformability of objects.
An ideal SLAM solution should be able to reason about dynamics in the environment including non-rigidity, work over long time periods generating ``all terrain'' maps and be able to do so in real time. 
In the computer vision community, there have been several attempts since the 80s to recover shape from non-rigid objects but with restrictive applicability. 
Recent results in non-rigid SfM such as~\cite{Garg2013,Grasa2014} are less restrictive but only work in small scenarios.
In the SLAM community, Newcombe\setal~\cite{newcombe2015dynamicfusion} have address the non-rigid case for small-scale reconstruction.
However, addressing the problem of non-rigid maps at a large scale is still largely unexplored. 

\mySubParagraph{Automatic parameter tuning}  SLAM systems (in particular, the data association modules) require extensive 
parameter tuning in order to work correctly for a given scenario. 
These parameters 
include thresholds that control feature matching, RANSAC parameters, 
 and criteria to
decide when to add new factors to the graph or when to trigger a loop closing algorithm to search for matches. 
If SLAM has to work ``out of the box" in arbitrary scenarios, methods for automatic tuning of the involved parameters need to be considered. 
\section{Long-term autonomy II: Scalability}
\label{sec:scalability}


While modern SLAM algorithms have been successfully demonstrated \blue{mostly} in indoor building-scale environments, 
in many application endeavors, robots must operate for an extended period of time over larger areas.
These applications include ocean exploration for environmental monitoring, non-stop cleaning robots in our \blue{ever changing} cities, or large-scale precision agriculture.
For such applications the size of the factor graph underlying SLAM can grow unbounded, due to the continuous exploration of new places and the increasing time of operation. 
In practice, the computational time and memory footprint are bounded by the resources of the robot.
Therefore, it is important to design SLAM methods whose computational and memory complexity remains bounded. 

In the worst-case, successive linearization methods based on \emph{direct} linear solvers imply a memory consumption which grows quadratically in the number of variables. 
When using iterative linear solvers (e.g., the conjugate gradient~\cite{Dellaert10iros}) the memory consumption grows linearly in the number of variables.
The situation is further complicated by the fact that, when re-visiting a place multiple times, factor graph optimization becomes less efficient as nodes and edges are continuously added to the same spatial region, compromising the sparsity structure of the graph.
 
In this section we review some of the current approaches to control, 
or at least reduce, the growth of the size of the problem and discuss open challenges.

\myParagraph{Brief Survey}
We focus on two ways to reduce the complexity of factor graph optimization: 
(i) \emph{sparsification methods}, which trade off information loss for memory and computational efficiency, 
and (ii) \emph{out-of-core and multi-robot methods}, which split the computation among many robots/processors.

\mySubParagraph{Node and edge sparsification}
This family of methods addresses scalability by reducing the number of nodes \emph{added} to the graph, 
or by \emph{pruning} less ``informative'' nodes and factors.
Ila\setal~\cite{ila-TRO10} use an information-theoretic approach to add only non-redundant nodes and highly-informative measurements to the graph. 
Johannsson\setal~\cite{johannsson-icra2013}, when possible, avoid adding new nodes to the graph by inducing new constraints between existing nodes, 
such that the number of variables grows only with size of the explored space and not with the mapping duration.
Kretzschmar\setal~\cite{KretzschmarIROS09} propose an information-based criterion for determining which nodes to marginalize in pose graph optimization.
Carlevaris-Bianco and Eustice~\cite{carlevaris-icra2013}, and Mazuran\setal~\cite{Mazuran-ijrr2016} introduce the \emph{Generic Linear Constraint} (GLC) factors and the \emph{Nonlinear Graph Sparsification} (NGS) method, respectively. These methods operate on the Markov blanket of a marginalized node and compute a sparse approximation of the blanket.
Huang\setal~\cite{huang-ecmr2013} sparsify the Hessian matrix 
\blue{(arising in the normal equations)} by 
solving an $\ell_1$-regularized minimization problem.

Another line of work that allows reducing the number of parameters to be estimated 
over time is the \emph{continuous-time trajectory estimation}. 
The first SLAM approach of this class was proposed by Bibby and Reid using cubic-splines to represent the continuous trajectory of the robot \cite{Bibby10icra}. 
In their approach the nodes in the factor graph represented the control-points (knots) of the spline which were optimized in a sliding window fashion. 
Later, Furgale\setal~\cite{Furgale12icra} proposed the use of basis functions, particularly B-splines, to approximate the robot trajectory, within a batch-optimization formulation. 
Sliding-window B-spline formulations were also used in SLAM with rolling shutter cameras, 
with a landmark-based representation by Patron-Perez\setal~\cite{patron2015spline} and with a semi-dense direct representation by Kim\setal~\cite{kim-icra16-rolling}. 
More recently, Mueggler\setal~\cite{Mueggler15rss} applied the continuous-time SLAM formulation to event-based cameras.
Bosse\setal~\cite{bosse2012zebedee} extended the continuous 3D scan-matching formulation from \cite{bosse2009continuous} to a large-scale SLAM application.
Later, Anderson\setal~\cite{anderson2014wavelet} and Dub\'e\setal~\cite{dube-icra16-nonuniform} proposed more efficient implementations by using wavelets or sampling non-uniform knots over the trajectory, respectively.
Tong\setal~\cite{tong2013gaussian} changed the parametrization of the trajectory from basis curves to a Gaussian process representation, where nodes in the factor graph are actual robot poses and any other pose can be interpolated by computing the posterior mean at the given time.
An expensive batch Gauss-Newton optimization is needed to solve for the states in this first proposal. 
Barfoot\setal~\cite{Barfoot-RSS-14} then proposed a Gaussian process with an exactly sparse inverse kernel that 
drastically reduces  the computational time of the batch solution. 

\mySubParagraph{Out-of-core (parallel) SLAM}
Parallel \emph{out-of-core} algorithms for SLAM split the computation (and memory) load of factor graph optimization among multiple processors.
The key idea is to divide the factor graph into different subgraphs and optimize the overall graph by alternating local optimization of each subgraph, with a global refinement.
The corresponding approaches are often referred to as \emph{submapping algorithms}, an idea that dates back to the initial attempts to tackle large-scale maps \cite{Bosse-IJRR04}. 
\blue{
Ni\setal~\cite{Ni10iros} and Zhao\setal~\cite{Zhao13iros} present submapping approaches for factor graph optimization, organizing the submaps in a binary tree structure. 
Grisetti\setal~\cite{Grisetti10icra} propose a hierarchy of submaps: whenever an observation is acquired, the highest level of the hierarchy is modified and only the areas which are substantially affected are changed at lower levels.
Some methods approximately decouple localization and mapping in two threads that run in parallel like Klein and Murray~\cite{KleinPTAM-07}. 
Other methods resort to solving different stages in parallel: inspired by \cite{Sibley-RSS09}, Strasdat\setal~\cite{strasdat-ICCV11} take a two-stage approach and optimize first a local pose-features graph and then a pose-pose graph; Williams\setal~\cite{Williams-ijrr2014} split factor graph optimization in a high-frequency filter and low-frequency smoother, 
which are periodically synchronized.
}

\mySubParagraph{Distributed multi robot SLAM}
One way of mapping a large-scale environment is to deploy multiple robots doing SLAM, and divide the scenario in smaller areas, 
each one mapped by a different robot.  
This approach has two main variants: the \emph{centralized} one, where robots build submaps and transfer the local information to a central station 
that performs inference~\cite{riazuelo-ras2014,Dong15icra}, 
and the \emph{decentralized} one, where there is no central data fusion and the agents leverage local communication to reach consensus on a common map.
Nerurkar\setal~\cite{Nerurkar09icra} propose an algorithm for cooperative localization based on distributed conjugate gradient. 
Araguez\setal~\cite{Aragues12tro} investigate consensus-based approaches for map merging.
Knuth and Barooah~\cite{Knuth13icra} estimate 3D poses using distributed gradient descent.
In Lazaro\setal~\cite{Lazaro13iros}, robots exchange portions of their factor graphs,
which are approximated in the form of condensed measurements to minimize communication. 
Cunnigham\setal~\cite{Cunningham13icra} 
use Gaussian elimination, and develop an approach, called DDF-SAM, in which each robot exchanges a Gaussian marginal over the~\emph{separators} (i.e., the variables shared by multiple robots).
A recent survey on multi-robot SLAM approaches can be found in \cite{saeedi-jfr2016}.

While Gaussian elimination has become a popular approach it has two major shortcomings. 
First, the marginals to be exchanged among the robots are dense, and the communication cost is quadratic in the number of separators. 
This motivated the use of sparsification techniques to reduce the communication cost~\cite{Paull15icra}.
The second reason is that Gaussian elimination is performed on a linearized version of the problem, hence approaches such as DDF-SAM~\cite{Cunningham13icra} require good linearization points and complex bookkeeping to ensure consistency of the linearization points across the robots. 
An alternative approach to Gaussian elimination is the Gauss-Seidel approach of Choudhary\setal~\cite{Choudhary16icra}, 
which implies a communication burden which is linear in the number of separators.

\myParagraph{Open Problems}
Despite the amount of work to reduce complexity of factor graph optimization, 
the literature has large gaps on other aspects related to long-term operation. 


\mySubParagraph{Map \YL{representation}} 
A fairly unexplored question is how to store the map during long-term operation. Even when memory is not a tight constraint, 
e.g. data is stored on the cloud, raw representations as point clouds or volumetric maps (see also Section~\ref{sec:rep-metric}) are wasteful in terms of memory; similarly, storing feature descriptors for vision-based SLAM quickly becomes cumbersome.
Some initial solutions have been recently proposed for localization against a compressed known map~\cite{Lynen-RSS-15},
 and for memory-efficient dense reconstruction~\cite{Klingensmith-RSS-15}.

\mySubParagraph{Learning, forgetting, remembering} 
A related open question for long-term mapping is how often to update the information contained in the map and how to decide when this information  becomes  outdated and can be discarded.
When is it fine, \YL{if ever}, to forget? \YL{In which case,} what can be forgotten and what is essential to maintain?
\YL{Can parts of the map be ``off-loaded'' and recalled when needed?}
While this is clearly task-dependent, no grounded answer to these questions 
has been proposed in the literature.


\mySubParagraph{Robust distributed mapping}
While approaches for outlier rejection have been proposed in the single robot case,  the 
literature on multi robot SLAM~\blue{barely deals with} the problem of outliers. 
Dealing with spurious measurements is particularly challenging for two reasons.
First, the robots might not share a common reference frame, making it harder to detect and reject wrong 
loop closures. Second, in the distributed setup, the robots have to detect outliers from very partial and local 
information. 
\blue{An early attempt to tackle this issue is~\cite{Fox2006}, in which robots actively verify location hypotheses using a rendezvous strategy before fusing information. Indelman\setal~\cite{Indelman16csm} propose a probabilistic approach to establish a common reference frame in the face of spurious measurements.}


\mySubParagraph{Resource-constrained platforms} 
Another relatively unexplored issue is how to adapt existing SLAM algorithms to the 
case in which the robotic platforms have \emph{severe} computational constraints. 
This problem is of great importance when the size of the platform is scaled down, e.g., mobile phones, micro 
aerial vehicles, or robotic insects~\cite{Wood15robotbee}. 
Many SLAM algorithms are too expensive to run on these platforms, 
and it would be desirable to have algorithms in which one can tune a ``knob'' that 
allows to gently trade off accuracy for computational cost.
Similar issues arise in the multi-robot setting: how can we guarantee reliable operation for multi robot teams when facing 
tight bandwidth constraints and  communication
dropout? The ``version control'' approach of Cieslewski\setal~\cite{cieslewski-icra2015} is a first study in this direction.


\section{Representation I: Metric Map Models} 
\label{sec:rep-metric}

\begin{figure}
\centering
\includegraphics[height=0.37\columnwidth]{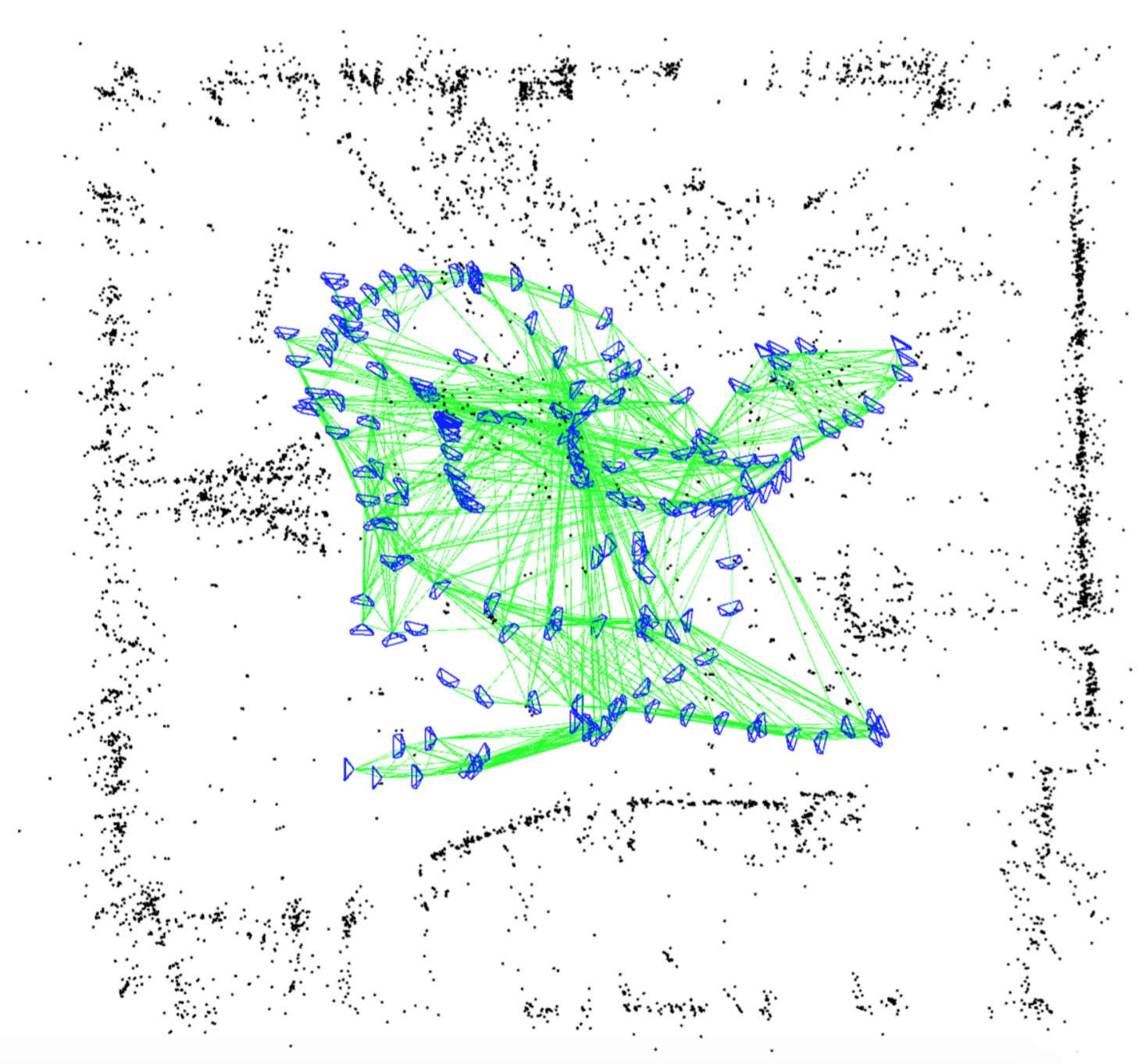} 
\hspace{3mm}
\includegraphics[height=0.37\columnwidth]{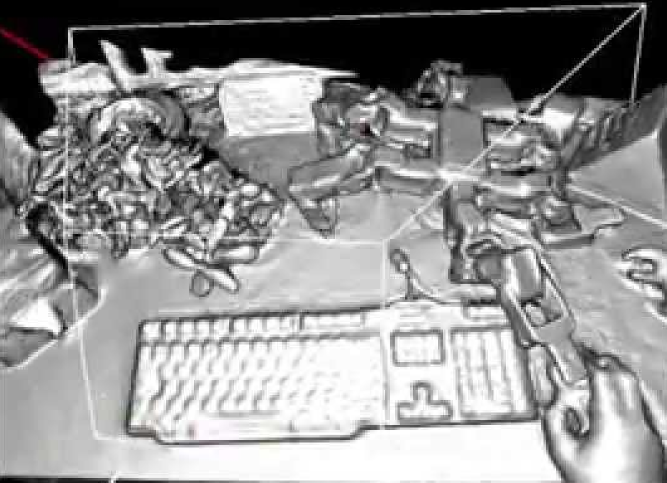}
\caption{Left: feature-based map of a room produced by
ORB-SLAM \cite{MurArtal15tro}. Right: dense map of a desktop produced by DTAM \cite{Newcombe11iccv}.}
\label{fig:sparse_dense}
\end{figure}

This section discusses how to model geometry in SLAM. 
More formally, a \emph{metric representation} (or metric map) is a symbolic structure that encodes 
the geometry of the environment. We claim that understanding 
 how to choose a suitable metric representation for SLAM (and extending the set or representations
currently used in robotics) will impact many research areas, including long-term 
navigation, physical interaction with the environment, and human-robot interaction. 

Geometric modeling appears much simpler in the 2D case, 
with only two predominant paradigms: \emph{landmark-based maps} and 
\emph{occupancy grid maps}. 
The former models the environment as a sparse set of landmarks, 
the latter discretizes the environment in cells and assigns a probability of occupation 
to each cell. The problem of standardization of these representations 
in the 2D case has been tackled by the
\emph{IEEE RAS Map Data Representation Working Group}, which recently 
released a standard for 2D maps in robotics~\cite{mapRepresentations}; the 
standard defines the two main metric representations for planar environments 
(plus topological maps) in order  
to facilitate data exchange, benchmarking, and technology transfer. 

The question of 3D geometry modeling is more delicate, and the 
understanding of how to efficiently model 3D geometry during mapping is 
in its infancy. 
In this section we review metric representations, 
taking a broad perspective across robotics, 
computer vision, 
computer aided design (CAD), and computer graphics.
Our taxonomy draws 
inspiration from~\cite{Foley92book,Shapiro02chapter,Requicha80csur}, and includes 
 pointers to more recent work. 


\myParagraph{Landmark-based sparse representations}
Most SLAM methods represent the scene as a set of \emph{sparse} 3D landmarks corresponding to discriminative features in the environment (e.g., lines, corners)~\cite{MurArtal15tro}; one example is shown in \prettyref{fig:sparse_dense}(left). These are commonly referred to 
as \emph{landmark-based} or \emph{feature-based} representations, and have been widespread in mobile robotics 
since early work on localization and mapping, 
and in computer vision in the context of
\emph{Structure from Motion}~\cite{Triggs00,Agarwal10eccv}. 
A common assumption underlying these representations is that the landmarks are distinguishable, 
i.e., sensor data measure some geometric aspect of the landmark, but also provide 
a \emph{descriptor} which establishes a (possibly uncertain) data association between each measurement 
and the corresponding landmark. 
\blue{Previous work also investigates different 3D landmark parameterizations, 
including global and local Cartesian models, and inverse depth parametrization~\cite{Montiel06rss}.}
 While a large body of work focuses on the estimation of point features, 
 the robotics literature includes extensions to  
  more complex geometric landmarks, including 
 lines, segments, or arcs~\cite{Lu15tro}. 

\myParagraph{Low-level raw dense representations}
Contrary to landmark-based representations, dense representations attempt to provide 
high-resolution models of the 3D geometry; these models are  more suitable for 
obstacle avoidance, or for visualization and rendering, see \prettyref{fig:sparse_dense}(right).
Among dense models, \emph{raw representations} describe the 3D geometry by means of a large 
unstructured set of points (i.e., \emph{point clouds}) or polygons
(i.e., \emph{polygon soup}~\cite{Shen04iai}). 
Point clouds have been widely used in robotics, in conjunction with 
 stereo and RGB-D cameras, as well as 3D laser scanners~\cite{Nuchter09book}.
These representations have recently gained popularity in monocular SLAM, in conjunction with 
the use of \emph{direct methods}~\cite{Irani99,Newcombe11iccv,Pizzoli14icra}, 
which  
estimate the trajectory of the robot and a 3D model directly from the intensity values of all the image pixels.
Slightly more complex representations are \emph{surfel maps}, which encode the geometry 
as a set of disks~\cite{Henry10iser,Whelan15rss}.
While these representations are visually pleasant, they are usually cumbersome as 
they require storing a large amount of data. Moreover, they give a low-level description of the 
geometry, neglecting, for instance, the topology of the obstacles.

\myParagraph{Boundary and spatial-partitioning dense representations}
These representations go beyond unstructured sets of low-level primitives (e.g., points) 
and attempt to explicitly represent surfaces (or \emph{boundaries}) and 
volumes. \blue{These representations lend themselves  better to tasks such as motion or footstep planning, obstacle avoidance, manipulation, and other physics-based reasoning, such as contact reasoning.}
Boundary representations (b-reps) define 3D
objects in terms of their surface boundary. 
Particularly simple boundary representations are plane-based models, which 
have been used for mapping by Castle\setal~\cite{Castle-ICRA2007} 
and Kaess~\cite{Kaess15icra,Lu15tro}.
More general b-reps include \emph{curve-based representations} (e.g., tensor product of NURBS or B-splines), 
  \emph{surface mesh models} (connected sets of polygons), and \emph{implicit surface 
  representations}. The latter specify the surface of a solid as the zero crossing
 of a function defined on $\Real{3}$~\cite{Bloomenthal97book};  
 examples of functions include \emph{radial-basis functions}~\cite{Carr01siggraph}, 
  \emph{signed-distance function}~\cite{Curless96siggraph}, and 
  \emph{truncated signed-distance function} (TSDF)~\cite{Zach07iccv}.
  \blue{TSDF are currently a popular representation for vision-based SLAM in robotics, attracting increasing attention 
  after the seminal work~\cite{Newcombe11ismar}.}
  Mesh models have been also used in~\cite{Whelan12rgbd,Whelan15rss}.

Spatial-partitioning representations define 3D objects
 as a collection of contiguous non-intersecting primitives. 
The most popular spatial-partitioning representation is
the so called \emph{spatial-occupancy enumeration}, which
 decomposes the 3D space into identical cubes (\emph{voxels}), arranged in a regular 3D grid. 
  More efficient partitioning schemes include 
  \emph{octree}, \emph{Polygonal Map octree}, and 
 \emph{Binary Space-Partitioning tree}~\cite[\S 12.6]{Foley92book}. 
 In robotics, octree representations have been used for 3D mapping~\cite{fairfield2007real}, 
 while commonly used \emph{occupancy grid maps}~\cite{Elfes87ra}
 can be considered as probabilistic variants of spatial-partitioning representations. 
 In 3D environments without hanging obstacles, 2.5D elevation maps have been also used~\cite{Brand14iros}.
Before moving to higher-level representations, let us better understand  how 
sparse (feature-based) representations (and algorithms) compare to dense ones in visual SLAM. 

\emph{Which one is best: feature-based or \blue{direct methods?}} 
%
Feature-based approaches are quite mature, \blue{with a long history of success}~\cite{Davison07pami}.
They allow to build accurate and robust SLAM systems with automatic relocation and loop closing~\cite{MurArtal15tro}.
However, such systems depend on the availability of features in the environment, the reliance on detection and matching thresholds, and on the fact that most feature detectors are optimized for speed rather than precision.
\blue{On the other hand, direct methods work with the raw pixel information and}
dense-direct methods exploit all the information in the image, even from areas where gradients are small; thus, they can outperform feature-based methods in scenes with poor 
texture, defocus, and motion blur \cite{Newcombe11iccv, Pizzoli14icra}. However, they require high computing power (GPUs) for real-time performance. 
Furthermore, how to jointly estimate dense structure and motion is still an open problem (currently they can be only be estimated subsequently to one another).
To avoid the caveats of feature-based methods there are two alternatives. 
\emph{Semi-dense} methods overcome the high-computation requirement of dense method by exploiting only pixels with strong gradients (i.e., edges) \cite{Engel14eccv, Forster16svo2};
\emph{semi-direct} methods instead leverage both sparse features (such as corners or edges) and direct 
methods \cite{Forster16svo2} and are proven to be the most efficient \cite{Forster16svo2}; 
additionally, because they rely on sparse features, they allow joint estimation of structure and motion.

\myParagraph{High-level object-based representations}
While point clouds and boundary representations are currently dominating 
the landscape of dense mapping, we envision that higher-level representations, 
including objects and solid shapes, will play a key role in the future of SLAM. 
Early techniques to include object-based reasoning in SLAM 
are ``$\text{SLAM++}$'' from~Salas-Moreno\setal~\cite{SalasMoreno13cvpr}, the work from Civera\setal~\cite{civera-iros2011}, and
 Dame\setal~\cite{dame-cvpr2013}.
Solid representations explicitly encode the fact that real objects are three-dimensional 
rather than 1D (i.e., points), or 2D (surfaces). 
Modeling objects as solid shapes 
allows associating physical notions, such as volume and mass, to each object, which 
is definitely important for robots which have to interact the world.
Luckily, existing literature from CAD and computer graphics paved the way towards these developments.
In the following, we list few examples of solid representations that have not yet been 
used in a SLAM context:

\begin{itemize}[leftmargin=*] 

\item  \mySubParagraph{Parameterized Primitive Instancing} relies on the definition of 
\emph{families} of objects (e.g., cylinder, sphere). For each family, one defines a set of parameters 
(e.g., radius, height), 
that uniquely identifies a member (or \emph{instance}) of the family. 
\blue{This representation may be of interest for SLAM since it enables the use of extremely compact 
models, while still capturing many elements in man-made environments.}
 \notAddressed{JN: Is ``parameterized primitive instancing'' really of interest?}

\item  \mySubParagraph{Sweep representations} define a solid as 
the sweep of a 2D or 3D object along a trajectory through space. 
Typical sweeps representations include translation sweep (or \emph{extrusion}) 
and rotation sweep. For instance, a cylinder can be represented as a translation sweep 
of a circle along an axis that is orthogonal to the plane of the circle.
Sweeps of 2D cross-sections are known as \emph{generalized cylinders} in computer vision~\cite{Binford71csc}, 
and they have been used in robotic grasping~\cite{Phillips15icra}. 
\blue{This representation seems particularly suitable to reason on the 
occluded portions of the scene, by leveraging symmetries.}

\item  \mySubParagraph{Constructive solid geometry} defines
complex solids by means of 
boolean operations between primitives~\cite{Requicha80csur}.
An object is stored as a tree in which the leaves are the primitives and the edges 
represent operations. 
\blue{This representation can model fairly complicated geometry and is extensively used 
in computer graphics.}
\end{itemize}

We conclude this review by mentioning that other types of representations exist, including 
feature-based models in CAD~\cite{Shah95book},
dictionary-based representations~\cite{Zhang12mia},
affordance-based models~\cite{Kim14siggraph}, 
generative and procedural models~\cite{Merrell10tovg}, and
scene graphs~\cite{Johnson15cvpr}. 
In particular, dictionary-based representations, which define a solid as a combination of 
atoms in a dictionary, have been considered in robotics and computer vision, 
with dictionary learned from data~\cite{Zhang12mia} 
or based on existing repositories of object models~\cite{Lai10ijrr,Lim13iccv}. 
\myParagraph{Open Problems} 
The following problems regarding metric representation for SLAM 
deserve a large amount of fundamental research, and are still vastly unexplored.

\mySubParagraph{High-level, expressive representations in SLAM}
While most of the robotics community is currently focusing on point clouds or 
TSDF to model 3D geometry, these representations have two main drawbacks. 
First, they are wasteful of memory. For instance, both representations use 
many parameters (i.e., points, voxels) to encode even a simple environment, such as an empty room \blue{(this 
issue can be partially mitigated by the
 so-called voxel hashing \cite{niessner2013hashing}).}
Second, these representations do not provide any high-level understanding of the 
3D geometry. For instance, consider the case in which the robot has to 
figure out if it is moving in a room or in a corridor. 
A point cloud does not provide readily usable information about the 
type of environment (i.e., room vs. corridor). 
On the other hand, more 
sophisticated models (e.g., parameterized primitive instancing) 
would provide easy ways to discern the two scenes (e.g., by looking at the parameters 
defining the primitive). 
Therefore, the use of higher-level representations in SLAM carries three promises.
First, using more compact representations would provide a natural tool for map compression in 
  large-scale mapping. 
  \blue{Second, high-level representations would provide a higher-level description of objects geometry which 
  is a desirable feature to facilitate data association, place recognition, semantic understanding, and human-robot interaction; 
  these representations would also provide a powerful support for SLAM, enabling to reason about occlusions, leverage shape priors, 
  and inform the inference/mapping process of the physical properties of the objects (e.g., weight, dynamics).}
Finally, using rich 3D representations would enable interactions with existing standards for 
construction and management of modern buildings, including CityGML~\cite{cityGML} and IndoorGML~\cite{IndoorGML}.
  No SLAM techniques can currently build higher-level 
 representations, beyond point clouds, mesh models, surfels models, and TSDFs.
 Recent efforts in this direction include~\cite{Cohen12cvpr,Srinivasan15bmvc,BodisSzomoru15cviu}.

\mySubParagraph{Optimal Representations} 
While there is a relatively large body of literature on different representations for 
3D geometry, few works have focused on understanding which criteria should guide the 
choice of a specific representation. 
Intuitively, in simple indoor environments one should prefer parametrized 
primitives since few parameters can sufficiently describe the 3D geometry; 
on the other hand, in complex outdoor environments, one might prefer mesh models.
Therefore, how should we compare different representations and 
how should we choose the ``optimal'' representation?
%
\notAddressed{JN: This sentence makes no sense:} Requicha~\cite{Requicha80csur} 
identifies few basic properties of solid representations that allow comparing
different representation. 
Among these properties we find: \emph{domain} (the set of real objects that can be represented), 
 \emph{conciseness} (the ``size'' of a representation for storage and transmission),
\emph{ease of creation} (in robotics this is the ``inference'' time required for the construction 
of the representation), and \emph{efficacy in the context of the application} 
(this depends on the tasks for which the representation is used).
Therefore, the ``optimal'' representation is the one that 
enables preforming a given task, while being concise and easy to create.
Soatto and Chiuso~\cite{Soatto14iclr} define the optimal representation as a minimal 
sufficient statistics to perform a given task, and its maximal invariance to nuisance factors.
%
%
Finding a general yet tractable framework to choose the best representation for a 
 task remains an open problem. 

\mySubParagraph{Automatic,  Adaptive Representations} 
Traditionally, the choice of a representation has been entrusted to the roboticist designing the system, but this has two main drawbacks. First, the design of a suitable representation is a time-consuming task
 that requires an expert. Second, it does not allow any flexibility: once the system is designed, the representation of choice cannot be changed; ideally, we would like a robot to use more or less complex representations 
 depending on the task and the complexity of the environment. 
  The automatic design of optimal representations will have a large impact on long-term navigation.
  
%
%
%
%




\section{Representation II: Semantic Map Models} 
\label{sec:rep-semantic}

Semantic mapping consists in associating semantic concepts to geometric entities in a robot's surroundings.
Recently, the limitations of purely geometric maps have been recognized 
and this has spawned a significant and ongoing body of work in semantic mapping of environments, in order to enhance robot's autonomy and robustness, facilitate more complex tasks (e.g. avoid muddy-road while driving), move from path-planning to task-planning, and enable advanced human-robot interaction
\cite{cadena2015icra,bao-cvpr12,SalasMoreno13cvpr}.
These observations have led to different approaches for semantic mapping which vary in the numbers and types of semantic concepts and means of associating them with different parts of the environments.
As an example, Pronobis and Jensfelt~\cite{pronobis-icra2012} label different rooms, while Pillai and Leonard~\cite{pillai-rss15} segment several known objects in the map.
With the exception of few approaches, semantic parsing at the basic level was formulated as a classification problem, where simple mapping between the sensory data and semantic concepts has been considered. 

\myParagraph{Semantic vs. topological SLAM} 
As mentioned in~Section~\ref{sec:intro}, topological mapping drops the metric information and only leverages 
place recognition to build a graph in which the nodes represent distinguishable ``places'', while edges 
denote reachability among places. 
We note that \emph{topological} mapping is radically different from \emph{semantic} mapping.
While the former requires recognizing a previously seen place (disregarding whether that place is a kitchen, a corridor, etc.), the latter is interested in classifying the place according to semantic labels.
A comprehensive survey on vision-based topological SLAM is presented in Lowry\setal~\cite{lowry-tro2016rohtua}, and some of its challenges are discussed in Section~\ref{sec:robustness}.   
In the rest of this section we focus on semantic mapping.

\myParagraph{Semantic SLAM: Structure and detail of concepts}
The unlimited number of, and relationships among, concepts for humans opens a more philosophical and task-driven decision about the level and organization of the semantic concepts.
The detail and organization depend on the context of what, and where, the robot is supposed to perform a task, and they impact the complexity of the problem at different stages. A semantic representation is built by defining the following aspects:
\begin{itemize}[leftmargin=*] 
\item \mySubParagraph{Level/Detail of semantic concepts}  For a given robotic task, e.g. ``going from room A to room B'', coarse categories (rooms, corridor, doors) would suffice for a successful performance, while for other tasks, e.g. ``pick up a tea cup'', finer categories (table, tea cup, glass) are needed.
\item \mySubParagraph{Organization of semantic concepts} The semantic concepts are not exclusive. Even more, a single entity can have an unlimited number of properties or concepts. A chair can be ``movable'' and ``sittable''; a dinner table can be ``movable'' and ``unsittable''. While the chair and the table are pieces of furniture,  they share the movable property but with different usability. Flat or hierarchical organizations, sharing or not some properties, have to be designed to handle this multiplicity of concepts.
\end{itemize}

\begin{figure}
\centering
\includegraphics[width=\columnwidth]{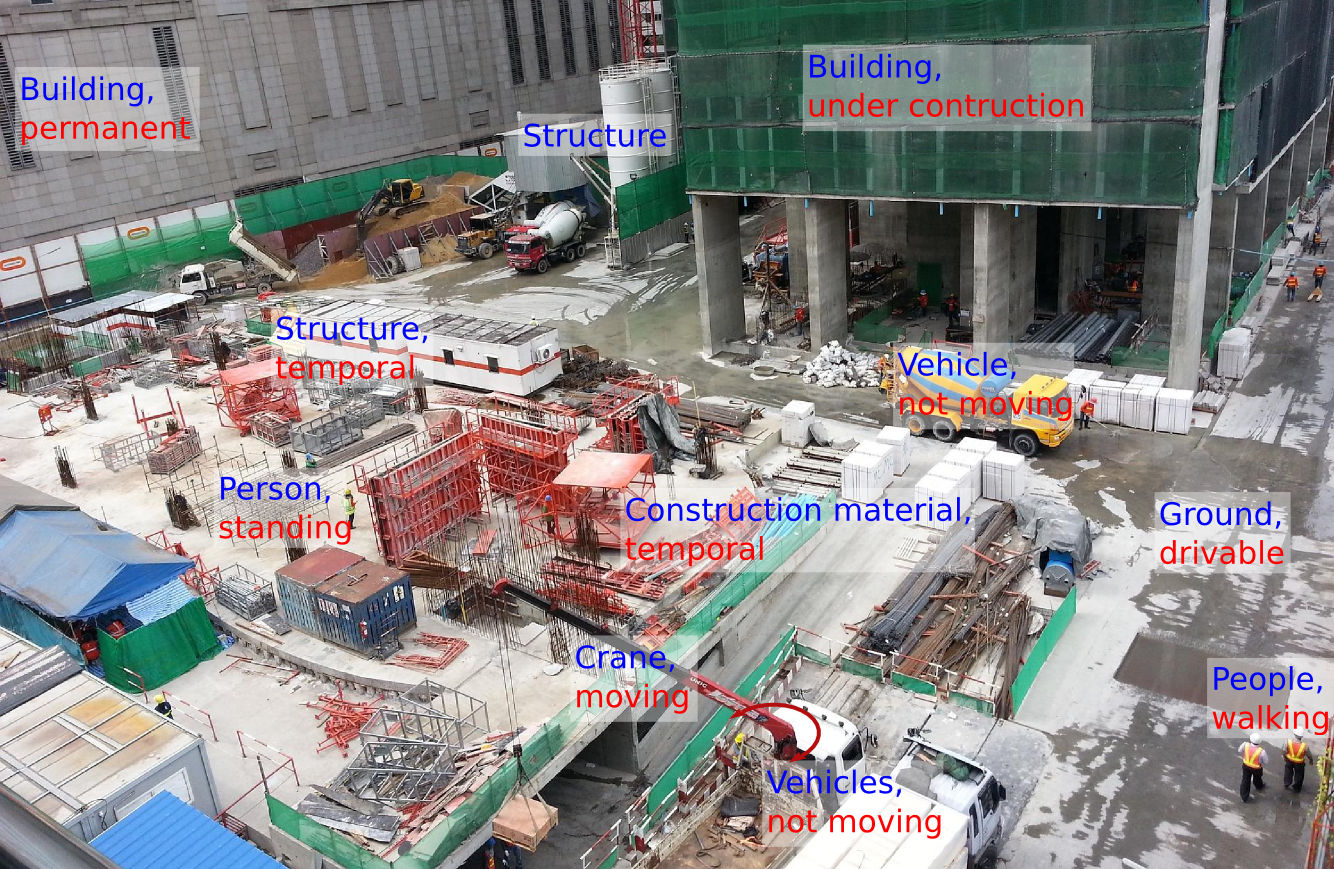}
\caption{
Semantic understanding allows humans to predict changes in the environment at different time scales. For instance, in the construction site shown in the figure, humans account for the motion of the crane and expect the crane-truck 
not to move in the immediate future, while at the same time we can predict the semblance of the site which will allow us to localize even after the construction finishes.
This is possible because 
we reason on the functional properties and interrelationships of the entities in the environment.
Enhancing our robots with similar capabilities is an open problem for semantic SLAM.
}
\label{fig:semantics}
\vspace{-0.5cm}
\end{figure}

\myParagraph{Brief Survey}
There are three main ways to attack semantic mapping, and assign semantic concepts to data.

\mySubParagraph{SLAM helps Semantics}
The first robotic researchers working on semantic mapping started by the straightforward approach of segmenting the metric map built by a classical SLAM system into semantic concepts. 
An early work was that of Mozos\setal~\cite{mozos-ras2007}, which builds a geometric map using a 2D laser scan and then fuses the classified semantic places from each robot pose through an associative Markov network in an offline manner. 
\blue{
Similarly, Lai\setal~\cite{Lai2014icra} build a 3D map from RGB-D sequences to then carry out an offline object classification.
}
An online semantic mapping system was later proposed by Pronobis\setal~\cite{pronobis-icra2012}, who combine three layers of reasoning (sensory, categorical, and place) to build a semantic map of the environment using laser and camera sensors. More recently, Cadena\setal~\cite{cadena2015icra} use motion estimation, and interconnect a coarse semantic segmentation with different object detectors to outperform the individual systems. 
Pillai and Leonard~\cite{pillai-rss15} use a monocular SLAM system to boost the performance in the task of object recognition in videos.

\mySubParagraph{Semantics helps SLAM}
Soon after the first semantic maps came out, another trend started by taking advantage of known semantic classes or objects. The idea is that if we can recognize objects or other elements in a map then we can use our prior knowledge about their geometry to improve the estimation of that map. First attempts were done in small scale by Castle\setal~\cite{Castle-ICRA2007} and by Civera\setal~\cite{civera-iros2011} with a monocular SLAM with sparse features, and by Dame\setal~\cite{dame-cvpr2013} with a dense map representation. Taking advantage of RGB-D sensors, Salas-Moreno\setal~\cite{SalasMoreno13cvpr} propose a SLAM system based on the detection of known objects in the environment.

\mySubParagraph{Joint SLAM and Semantics inference}
Researchers with expertise in both computer vision and robotics realized that they could perform monocular SLAM and map segmentation within a joint formulation. 
The online system of Flint\setal~\cite{flint2011manhattan} presents a model that leverages the Manhattan world assumption to segment the map in the main planes in indoor scenes.
Bao\setal~\cite{bao-cvpr12} propose one of the first approaches to jointly estimate camera parameters, scene points and object labels using both geometric and semantic attributes in the scene. 
In their work, the authors demonstrate the improved object recognition performance and robustness, at the cost of a run-time of 20 minutes per image-pair, and the limited number of object categories makes the approach impractical for on-line robot operation.
In the same line, H{\"a}ne\setal~\cite{hane2013joint} solve a more specialized class-dependent optimization problem in outdoors scenarios.
Although still offline, Kundu\setal~\cite{kundu-eccv2014} reduce the complexity of the problem by a late fusion of the semantic segmentation and the metric map, a similar idea was proposed earlier by Sengupta\setal~\cite{sengupta-icra2015} using stereo cameras. It should be noted that \cite{kundu-eccv2014} and \cite{sengupta-icra2015} focus only on the mapping part and they do not refine the early computed poses in this late stage. 
Recently, a promising online system was proposed by Vineet\setal~\cite{vineet-icra2015} using stereo cameras and a dense map representation.

\myParagraph{Open Problems}
The problem of including semantic information in SLAM is in its infancy, and, contrary to metric SLAM, 
it still lacks a cohesive formulation. \prettyref{fig:semantics} shows a construction site as a simple example where we can find the challenges discussed below.

\blue{ 
\mySubParagraph{Consistent semantic-metric fusion}
Although some progress has been done in terms of temporal fusion of, for instance, per frame semantic evidence~\cite{vineet-icra2015,sengupta-icra2015}, the problem of consistently fusing several sources of semantic information with metric information coming at different points in time is still open.
Incorporating the confidence or uncertainty of the semantic categorization in the already well known factor graph formulation for the metric representation is a possible way to go for a joint semantic-metric inference framework.
}

\mySubParagraph{Semantic mapping is much more than a categorization problem} The semantic concepts are evolving to more specialized information such as affordances and actionability\footnote{The term \emph{affordances} refers to the set of possible actions on a given object/environment by a given agent \cite{gibson2014ecological}, while the term \emph{actionability} includes the expected utility of these actions.} of the entities in the map and the possible interactions among different active agents in the environment. How to represent these properties, and interrelationships, are questions to answer for high level human-robot interaction.

\mySubParagraph{Ignorance, awareness, and adaptation}
Given some prior knowledge, the robot should be able to reason about new concepts and their semantic representations, that is, it should be able to discover new objects or classes in the environment, learning new properties as result of active interaction with other robots and humans, and adapting the representations to slow and abrupt changes in the environment over time.
For example, suppose that a wheeled-robot needs to classify whether a terrain is drivable or not, to inform its navigation system.
If the robot finds some mud on a road, that was previously classified as drivable, the robot should learn a new class depending on the grade of difficulty of crossing the muddy region, or adjust its classifier if another vehicle stuck in the mud is perceived.

\mySubParagraph{Semantic-based reasoning\footnote{Reasoning in the sense of localization and mapping. This is only a sub-area of the vast area of \emph{Knowledge Representation and Reasoning} in the field of Artificial Intelligence that deals with solving complex problems, like having a dialogue in natural language or inferring a person's mood.}} 
As humans, the semantic representations allow us to compress and speed-up reasoning about the environment, while assessing accurate metric representations takes us some effort.
Currently, this is not the case for robots.
Robots can handle (colored) metric representation but they do not truly exploit the semantic concepts.
Our robots are currently unable to effectively, and efficiently localize and continuously map using the semantic concepts (categories, relationships and properties) in the environment.
For instance, when detecting a car, a robot should infer the presence of a planar ground  under the car (even if occluded) and when the car moves the map update should only refine the hallucinated ground with the new sensor readings.
Even more, the same update should change the global pose of the car as a whole in a single and efficient operation as opposed to update, for instance, every single voxel.


\section{New theoretical tools for SLAM}
\label{sec:theory}

This section discusses recent progress towards establishing performance guarantees for SLAM algorithms, 
and elucidates open problems. 
The theoretical analysis is important for three main reasons. 
First, SLAM algorithms and implementations are often tested in few problem instances and it is hard 
to understand how the corresponding results generalize to new instances.
Second, theoretical results shed light on the intrinsic properties of the problem, 
revealing aspects that may be counter-intuitive 
during empirical evaluation. 
Third, a true understanding of the structure of the problem allows pushing the algorithmic boundaries, 
enabling to extend the set of real-world SLAM instances that can be solved.

\blue{Early theoretical analysis of SLAM algorithms were based on 
the use of EKF; we refer the reader 
to~\cite{Huang11book,dissanayake-review2011} for a comprehensive discussion, on consistency and 
 observability of EKF SLAM.}\footnote{\blue{Interestingly, the 
lack of observability manifests itself very clearly in factor graph optimization, since the linear system to be 
solved in iterative methods becomes rank-deficient; this enables the design of techniques that can explicitly deal with problems that are not fully observable~\cite{Zhang16icra-degeneracy}}.}
Here we focus on factor graph optimization approaches. 
Besides the practical advantages (accuracy, efficiency), factor graph optimization 
provides an elegant framework which is more amenable  to analysis. 
In the absence of priors, MAP estimation reduces to maximum likelihood estimation. 
Consequently, without priors, SLAM inherits all the properties of maximum likelihood 
estimators: the estimator in \eqref{NonLinearLS} is consistent, asymptotically Gaussian, asymptotically 
efficient, and invariant to transformations in the Euclidean space~\cite[Theorems 11-1,2]{Mendel95book}.
Some of these properties are lost in presence of priors (e.g., the estimator is no 
longer invariant~\cite[page 193]{Mendel95book}). 

In this context we are more interested in \emph{algorithmic properties}: does a given algorithm converge 
to the MAP estimate? How can we improve or check convergence? What is the breakdown point 
in presence of spurious measurements?

\begin{figure}
\centering
\includegraphics[width=\columnwidth]{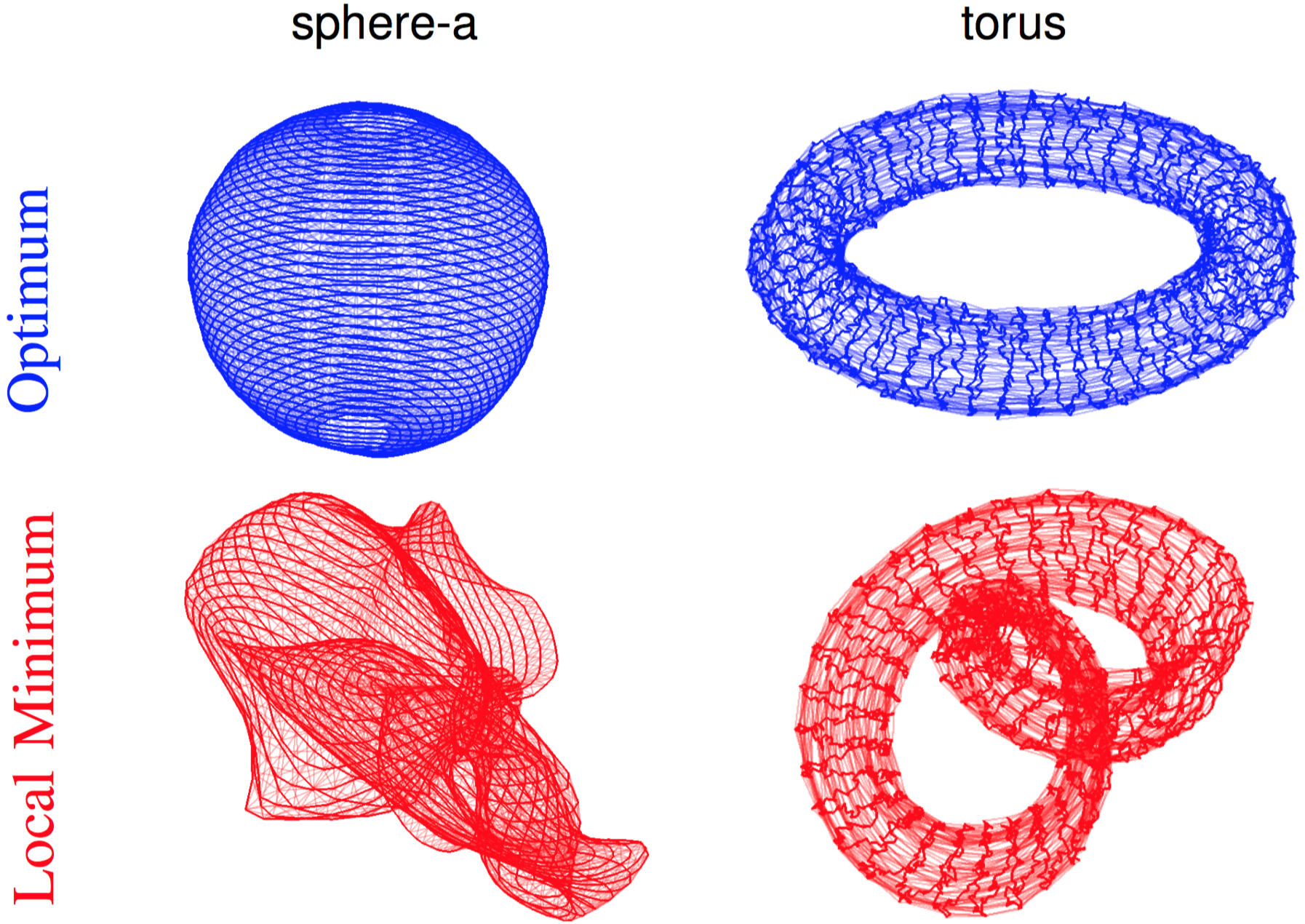}
\caption{The back-bone of most SLAM algorithms is the MAP estimation of the robot trajectory, which 
is computed via non-convex optimization. 
\blue{The figure shows trajectory estimates for two simulated benchmarking problems, namely
{\tt sphere-a} and {\tt torus}, in which the robot travels on the surface of a sphere and a torus. 
The top row reports the correct trajectory estimate, corresponding to the global optimum 
of the optimization problem. The bottom row shows incorrect trajectory estimates resulting 
from convergence to local minima.}
Recent theoretical tools are enabling detection of wrong convergence episodes, and are opening avenues for failure detection and recovery techniques. 
}
\label{fig:verification}
\end{figure}

\myParagraph{Brief Survey}
Most SLAM algorithms are based on iterative nonlinear optimization~\cite{Grisetti09its,Olson06icra,Dellaert-IJRR06,Kaess08tro,Kaess-ijrr2012,Polok13rss}.
SLAM is a nonconvex problem and iterative optimization can only guarantee local convergence. 
When an algorithm converges to a local minimum\footnote{We use the term ``local minimum'' to denote 
a minimum of the cost which does not attain the globally optimal objective.}
 it usually returns an estimate that is completely wrong and unsuitable for navigation (\blue{\prettyref{fig:verification}}).
State-of-the-art iterative solvers fail to converge to a global minimum of the cost for 
relatively small noise levels~\cite{Carlone14tro,Carlone15icra-init3D}.

Failure to converge in iterative methods has
 triggered efforts towards a deeper understanding of the SLAM problem.
Huang and collaborators~\cite{Huang10iros} pioneered this effort, with initial works discussing the 
nature of the nonconvexity in SLAM. 
Huang\setal~\cite{Huang12icra} 
discuss the number of minima in small pose graph optimization problems. 
Knuth and Barooah~\cite{Knuth13ras} investigate 
the growth of the error in the absence of loop closures. Carlone~\cite{Carlone13icra} 
provides estimates of the basin of convergence for the Gauss-Newton method. 
Carlone and Censi~\cite{Carlone14tro} show that rotation estimation can be solved 
in closed form in 2D and show that the corresponding estimate is unique.
The recent use of alternative maximum likelihood formulations (e.g., assuming Von Mises  
noise on rotations~\cite{Carlone15icra-verify,Rosen15icra}) has enabled even stronger results. 
Carlone and Dellaert~\cite{Carlone16tro-duality2D,Carlone15iros-duality3D} show that under certain conditions (strong duality)
that are often encountered in practice, 
the maximum likelihood estimate is unique and pose graph optimization can be solved globally, via (convex) semidefinite programming (SDP).
A very recent overview on theoretical aspects of SLAM is given in~\cite{huang2016ijars}.

As mentioned earlier, the theoretical analysis is sometimes the first step towards the design of better 
algorithms. Besides the dual SDP approach of~\cite{Carlone16tro-duality2D,Carlone15iros-duality3D}, 
other authors proposed convex relaxation to avoid convergence to local minima. These contributions 
include the work of Liu\setal~\cite{Liu12iros} and Rosen\setal~\cite{Rosen15icra}.
Another successful strategy to improve convergence consists in computing a suitable \emph{initialization} 
for iterative nonlinear optimization. In this regard, the idea of solving for the rotations first and 
to use the resulting estimate to bootstrap nonlinear iteration has been demonstrated to be very 
effective in practice~\cite{Bosse09ras,Carlone14ijrr,Carlone14tro,Carlone15icra-init3D}. 
Khosoussi\setal~\cite{Khosoussi15rss} leverage the (approximate) separability between translation and rotation to speed up optimization.

Recent theoretical results on the use of Lagrangian duality in SLAM 
also enabled the design of \emph{verification techniques}: 
given a SLAM estimate these techniques are able to judge 
whether such estimate is optimal or not. 
Being able to ascertain the quality of a given SLAM solution is crucial 
to design failure detection and recovery strategies for safety-critical applications.
The literature on verification techniques for SLAM is very recent:
current approaches~\cite{Carlone16tro-duality2D,Carlone15iros-duality3D} are able to perform verification by solving a sparse linear system and are guaranteed to provide a correct answer as long as strong duality holds (more on this point later).

We note that these results, proposed in a robotics context, provide 
a useful complement to related work in other communities, including
localization in multi agent systems~\cite{Chiuso08cis,Wang13ima,Tron12cdc,Piovan13automatica,Peters15sicon}, 
structure from motion in computer vision~\cite{Martinec07cvpr,Govindu01cvpr,Fredriksson12accv,Hartley13ijcv},
and cryo-electron microscopy~\cite{Singer10achm,Singer11siam}.

\myParagraph{Open Problems} Despite the unprecedented progress of the last years, 
 several theoretical questions remain open. 

 \mySubParagraph{Generality, guarantees, verification}
The first question regards the generality of the available results. 
Most results on guaranteed global solutions and verification techniques have been proposed in the 
context of pose graph optimization. Can these results be generalized to arbitrary factor graphs?
\blue{Moreover, most theoretical results assume the measurement noise to be isotropic or at least to 
be structured. Can we generalize existing results to arbitrary noise models?} 

 \emph{Weak or Strong duality?}
The works~\cite{Carlone16tro-duality2D,Carlone15iros-duality3D} 
show that when strong duality holds SLAM can be solved globally; moreover, they provide 
empirical evidence that strong duality holds in most problem instances encountered in practical 
applications. The outstanding problem consists in establishing \emph{a priori} conditions under which 
strong duality holds. We would like to answer the question ``given a set of sensors 
(and the corresponding measurement noise statistics) and a factor graph structure, does strong duality hold?''. 
The capability to answer 
this question would  define the domain of applications in which 
we can compute (or verify) global solutions to SLAM. 
This theoretical investigation would also provide fundamental insights in 
sensor design and active SLAM (Section~\ref{sec:ActiveSLAM}).

\mySubParagraph{Resilience to outliers}
The third question regards estimation in the presence of spurious measurements. 
While recent results provide strong guarantees for pose graph optimization, 
no result of this kind applies in the presence of outliers. Despite the work on 
robust SLAM (Section~\ref{sec:robustness}) and new modeling tools for the non-Gaussian noise case~\cite{Rosen2013}, 
the design of global techniques that are resilient to outliers and the 
design of verification techniques that can certify the correctness of a given estimate
in presence of outliers remain open.

\section{Active SLAM}
\label{sec:ActiveSLAM}

So far we described SLAM as an estimation problem that is carried out passively by the robot, 
\ie~the robot performs SLAM \emph{given} the sensor data, but without acting deliberately 
to collect it.
In this section we discuss how to leverage a robot's motion to 
 improve the mapping and localization results. 



The problem of controlling robot's motion in order to 
minimize the uncertainty of its map representation and localization
 is usually named \emph{active SLAM}. %
This definition stems from the well-known Bajcsy's active perception~\cite{Bajcsy1986} and Thrun's robotic exploration~\cite[Ch. 17]{Thrun2005} paradigms. 

\myParagraph{Brief Survey}
The first proposal and implementation of an active SLAM algorithm can be traced back to Feder~\cite{Feder1999a} 
while the name was coined in~\cite{Leung2006}. 
However, active SLAM has its roots in  ideas from artificial intelligence and robotic exploration that can be traced back to the early eighties~(\cf~\cite{Barto1981}). Thrun in~\cite{Thrun1995} concluded that solving the \textit{exploration-exploitation} dilemma, 
i.e., finding a balance between visiting new places (exploration) and reducing the uncertainty by re-visiting known 
areas (exploitation), provides a more efficient alternative with respect to random exploration or pure exploitation.



Active SLAM is a decision making problem and there are several general frameworks for decision making that can be used as backbone for exploration-exploitation decisions. One of these frameworks is the Theory of Optimal Experimental Design (TOED)~\cite{Pazman1986}~which, applied to active SLAM~\cite{Carrillo2012a,Carrillo2012b}, allows selecting future robot action based on the predicted map uncertainty. Information theoretic~\cite{Renyi1960,MacKay2002} approaches have been also applied to active SLAM~\cite{Stachniss2009,Carrillo2015a}; in this case decision making is usually guided by the notion of \emph{information gain}. Control theoretic approaches for active SLAM include the use of Model Predictive Control~\cite{Leung2006,Leung2006a}. A different body of works formulates active SLAM under the formalism of Partially Observably Markov Decision Process (POMDP)~\cite{Kaelbling1998}, which in general is known to be computationally intractable;
approximate but tractable solutions for active SLAM include Bayesian Optimization~\cite{Martinez2009} or efficient 
Gaussian beliefs propagation~\cite{Patil2015}, among others. 

A popular framework for active SLAM consists of selecting the best future action among a finite set 
of alternatives. This family of active SLAM algorithms proceeds in three main steps~\cite{Blanco2008,Carlone2014}: 
1) The robot identifies possible locations to explore or exploit, \ie~vantage locations, in its current estimate of the map; 2) The robot computes the utility of visiting each vantage point and selects the action with the highest utility; and 3) The robot carries out the selected action and decides if it is necessary to continue or to terminate the task. In the following, we discuss each point in details.


\mySubParagraph{Selecting vantage points}
Ideally, a robot executing an active SLAM algorithm should evaluate every possible action in the robot and map space, but the computational complexity of the evaluation grows exponentially with the search space which proves to be computationally intractable in real applications~\cite{Burgard2005,Martinez2009}. 
In practice, a small subset of locations in the map is selected, 
using techniques such as frontier-based exploration~\cite{Yamauchi1997,Keidar2014}. 
Recent works~\cite{VandenBerg2012} and~\cite{Indelman2015} have proposed approaches for continuous-space planning under uncertainty that can be used for active SLAM; currently these approaches can only guarantee convergence to locally optimal policies.
Another recent continuous-domain avenue for active SLAM algorithms is the use of potential fields. Some examples are \cite{Vallve2015}, which uses convolution techniques to compute entropy and select the robot's actions, and \cite{Jorge2015}, which resorts to 
the solution of a boundary value problem. 

\mySubParagraph{Computing the utility of an action}
Ideally, to compute the utility of a given action the robot should reason about the evolution of the 
posterior over the robot pose and the map, taking into account future (controllable) actions and future (unknown) measurements. 
If such posterior were known, an information-theoretic function, as the information gain, could be used to rank
the different actions~\cite{Bourgault2002,Stachniss2005}.
However, computing this joint probability analytically is, in general, computationally intractable~\cite{Stachniss2005,Fairfield2010,Carlone2014}. In practice, one resorts to approximations. 
Initial work considered the uncertainty of the map and the robot to be independent~\cite{Valencia2012} or conditionally independent~\cite{Stachniss2005}. Most of these approaches define the utility as a linear combination of metrics that quantify robot and map uncertainties~\cite{Bourgault2002,Carlone2014}. One drawback of this approach is that the scale of the numerical values of the two uncertainties is not comparable,~\ie~the map uncertainty is often orders of magnitude larger than the robot one, so manual tuning is required to correct it. Approaches to tackle this issue have been proposed 
for particle-filter-based SLAM~\cite{Carlone2014}, and for pose graph optimization~\cite{Carrillo2015a}.

The Theory of Optimal Experimental Design (TOED)~\cite{Pazman1986} can also be used to account for the utility of performing an action. In the TOED, every action 
is considered as a stochastic design, 
and the comparison among designs is done using their associated covariance matrices via the so-called optimality criteria, \eg~\aopt, \dopt~and \eopt. A study about the usage of optimality criteria in active SLAM can be found in~\cite{Carrillo2012a,Carrillo2015b}. 

\mySubParagraph{Executing actions or terminating exploration} 
While executing an action is usually an easy task, using well-established techniques from motion planning, 
the decision on whether or not the exploration task is complete, is currently an open challenge that we discuss in the following 
paragraph. 

  \myParagraph{Open Problems}
Several problems still need to be addressed, for active SLAM to have
impact in real applications. 

\mySubParagraph{Fast and accurate predictions of future states}
In active SLAM each action of the robot should contribute to 
reduce the uncertainty in the map and improve the localization accuracy; for this purpose, the robot must be able to forecast the effect of future actions on the map and robot’s localization. The forecast has to be fast to 
meet latency constraints and precise to effectively support the decision process.
In the SLAM community it is well known that loop closings are important to reduce uncertainty and to improve localization and mapping accuracy. Nonetheless, efficient methods for forecasting the occurrence and the effect of a loop closing are yet to be devised. 
Moreover, 
predicting the effects of future actions is still
a computational expensive task~\cite{Indelman2015}.
Recent approaches to forecasting future robot states can be found in the machine learning literature, and involve the use of spectral techniques~\cite{Song2010} and deep learning~\cite{Wahlstrom2015}.

\emph{Enough is enough: When do you stop doing active SLAM?} 
\blue{Active SLAM is a computationally expensive task: therefore a natural question is when we can stop doing active SLAM and switch to classical (passive) SLAM in order to focus resources on other tasks}.
Balancing active SLAM decisions and exogenous tasks is critical, 
since in most real-world tasks, active SLAM is only a means to achieve an intended goal.
Additionally, having a stopping criteria is a necessity because at some point it is provable that more information would lead not only to a diminishing return effect but also, in case of contradictory information, to an unrecoverable state (\eg~several wrong loop closures). Uncertainty metrics from TOED, which are task oriented, seem promising as stopping criteria, compared to information-theoretic metrics which are difficult to compare across systems~\cite{Carrillo2013}.

\mySubParagraph{Performance guarantees} Another important avenue is to look for mathematical guarantees for active SLAM and for
 near-optimal policies.
Since solving the problem exactly is intractable, it is desirable to have approximation algorithms with clear performance bounds. 
Examples of this kind of effort is the use of submodularity~\cite{Golovin2011} in the related field of active sensors placement.




\section{New Frontiers: Sensors and Learning}
\label{sec:newTrends}

The development of new sensors and the use of new computational tools have often been key drivers for SLAM.  
Section~\ref{sec:sensors} reviews unconventional and new sensors, as well as the challenges and opportunities  they pose in the context of SLAM.
Section~\ref{sec:deepLearning} discusses the role of (deep) learning as an important frontier for SLAM,  
analyzing the possible ways in which this tool is going to improve, affect, or even restate, the SLAM problem.


\subsection{New and Unconventional Sensors for SLAM}
\label{sec:sensors}


\toAdd{please mention
\begin{itemize}
\item light-field cameras
\item unconventional sensing (contact sensing)
\item industry perspective (Google's Soli) 
\end{itemize}
}

Besides the development of new algorithms, progress in SLAM (and mobile robotics in general)
has often been  triggered by the availability of novel sensors.
For instance, the introduction of 2D laser range finders enabled the creation of very robust SLAM systems, 
while 3D lidars have been a main thrust behind recent applications, such as autonomous cars. 
In the last ten years, a large amount of research has been devoted to vision sensors, 
with successful applications in augmented reality and vision-based navigation.

\blue{Sensing in robotics has been mostly dominated by lidars and conventional vision sensors.}
However, there are many alternative sensors that can be leveraged for SLAM, such as depth, light-field, and event-based cameras, which are now becoming a commodity hardware, as well as magnetic, olfaction, and thermal sensors.

\myParagraph{Brief Survey} We review the most relevant new sensors and their applications for SLAM, 
postponing a discussion on open problems to the end of this section. 

\blue{
\mySubParagraph{Range cameras}
Light-emitting depth cameras are not new sensors, but they became commodity hardware in 2010 with the advent the Microsoft Kinect game console.
They operate according to different principles, such as 
structured light, time of flight, interferometry, or coded aperture.
Structure-light cameras work by triangulation; thus, their accuracy is 
limited by the distance between the cameras and the pattern projector (structured light).
By contrast, the accuracy of Time-of-Flight (ToF) cameras only depends on the time-of-flight measurement device; 
thus, they provide the highest range accuracy (sub millimeter at several meters).
ToF cameras  became commercially available for civil applications around the year 2000 but only began to be used in mobile robotics in 2004~\cite{weingarten2004state}.
While the first generation of ToF and structured-light cameras was characterized by low signal-to-noise ratio and high price, 
they soon became popular for video-game applications, which contributed to making them affordable and improving their accuracy.
Since range cameras carry their own light source, they also work in dark and untextured scenes, which enabled the achievement of remarkable SLAM results~\cite{Newcombe11ismar}.
}

\blue{
\mySubParagraph{Light-field cameras}
Contrary to standard cameras, which only record the light intensity hitting each pixel, a light-field camera (also known as plenoptic camera), 
records both the intensity and the direction of light rays~\cite{NgStanford_2005}.
One popular type of light-field camera uses an array of micro lenses placed in front of a conventional image sensor to sense intensity, color, and directional information. 
Because of the manufacturing cost, commercially available light-field cameras still 
have relatively low resolution ($<$ 1MP), which is being overcome by current technological effort. 
Light-field cameras offer several advantages over standard cameras, such as depth estimation, noise reduction~\cite{dansereau2013light}, 
video stabilization~\cite{smith2009light}, isolation of distractors~\cite{DansereauCVIU2016}, and specularity removal~\cite{jachnik2012real}. 
Their optics also offers wide aperture and wide depth of field compared with conventional cameras~\cite{bishop2012light}.
}

\blue{
\mySubParagraph{Event-based cameras}
Contrarily to standard frame-based cameras, which send entire images at fixed frame rates, event-based cameras, such as the Dynamic Vision Sensor (DVS)~\cite{Lichtsteiner08ssc} or the Asynchronous Time-based Image Sensor (ATIS)~\cite{posch2008asynchronous}, 
only send the local pixel-level changes caused by movement in a scene at the time they occur. 

%
%
They have five key advantages compared to conventional frame-based cameras: a temporal latency of 1ms, an update rate of up to 1MHz, a dynamic range of up to 140dB (vs 60-70dB of standard cameras), a power consumption of 20mW (vs 1.5W of standard cameras), and very low bandwidth and storage requirements (because only intensity changes are transmitted).
These properties enable the design of a new class of SLAM algorithms that can operate in scenes characterized by high-speed motion~\cite{gallegoArxiv2016} and high-dynamic range~\cite{kim2016eccv,RAL_Rebecq_2016}, where standard cameras fail.
However, since the output is composed of a sequence of asynchronous events, traditional frame-based computer-vision algorithms are not applicable. 
This requires a \emph{paradigm shift} from the traditional computer vision approaches developed over the last fifty years.
Event-based 
real-time localization and mapping algorithms have recently been proposed~\cite{kim2016eccv,RAL_Rebecq_2016}. 
The design goal of such algorithms is that each incoming event can asynchronously change the estimated state of the system, thus, preserving the event-based nature of
the sensor and allowing the design of microsecond-latency control algorithms \cite{mueller2015low}.
%
}



\myParagraph{Open Problems} 
The main bottleneck of active range cameras is the maximum range and interference with other external light sources (such as sun light); 
however, these weaknesses can be improved by emitting more light power.

\blue{
Light-field cameras have been rarely used in SLAM because they are usually thought to increase the amount of data produced and require more computational power. However, recent studies have shown that they are particularly suitable for SLAM applications because they allow formulating the motion estimation problem as a linear optimization 
and can provide more accurate motion estimates if designed properly~\cite{DongIJRR_2013}.

Event-based cameras are revolutionary image sensors that overcome the limitations of standard cameras in scenes characterized by high dynamic range and high speed motion.
Open problems concern a full characterization of the sensor noise and sensor non idealities:  
event-based cameras have a complicated analog circuitry, with nonlinearities and biases that can change the sensitivity
of the pixels, and other dynamic properties, which make the events susceptible to noise.
Since a single event does not carry enough information for state estimation and because an event camera generate on average $100,000$ events a second, it can become intractable to do SLAM at the discrete times of the single events due to the rapidly growing size of the state space. 
Using a continuous-time framework~\cite{Bibby10icra}, the estimated trajectory can be approximated by a smooth curve in the space of rigid-body motions using basis functions 
(e.g., cubic splines), and optimized according to the observed events \cite{Mueggler15rss}.
While the temporal resolution is very high, the spatial resolution of event-based cameras is relatively low (QVGA), which is being overcome by current technological effort~\cite{Brandli2014iscas}.
Newly developed event sensors 
 overcome some of the original limitations: an ATIS sensor sends the magnitude of the pixel-level brightness; a DAVIS sensor \cite{Brandli2014iscas} can output both frames and events (this is made possible by embedding a standard frame-based sensor and a DVS into the same pixel array). This will allow tracking features and motion in the blind time between frames~\cite{Kueng_IROS_2016}.

We conclude this section with some general observations on 
the use of novel sensing modalities for SLAM.
}

\blue{
\mySubParagraph{Other sensors}
Most SLAM research has been devoted to range and vision sensors.
However, humans or animals are able to improve their sensing capabilities by using tactile, olfaction, sound, magnetic, and thermal stimuli.
For instance, tactile cues are used by blind people or rodents for haptic exploration of objects, olfaction is used by bees to find their way home, magnetic fields are used by homing pigeons for navigation, sound is used by bats for obstacle detection and navigation, while some snakes can see infrared radiation emitted by hot objects.
Unfortunately, these alternative sensors have not been considered in the same depth as range and vision sensors to perform SLAM.
Haptic SLAM can be used for tactile exploration of an object or of a scene~\cite{Strub_2014,Yu_IROS_15}. 
Olfaction sensors can be used to localize gas or other odor sources~\cite{marques2002olfaction}.
Although ultrasound-based localization was predominant in early mobile robots, their use has rapidly declined with the advent of cheap optical range sensors. 
Nevertheless, animals, such as bats, can navigate at very high speeds using only echo localization. 
Thermal sensors offer important cues at night and in adverse weather conditions~\cite{magnabosco2013cross}. 
Local anomalies of the ambient magnetic field, present in many indoor environments, offer an excellent cue for localization~\cite{vallivaara2010simultaneous}.
Finally, pre-existing wireless networks, such as WiFi, can be used to improve robot navigation without any prior knowledge of the location of the antennas~\cite{ferris2007wifi}.
}

\emph{Which sensor is best for SLAM?}
A question that naturally arises is: what will be the next sensor technology to drive future long-term SLAM research? 
Clearly, the performance of a given algorithm-sensor pair for SLAM depends on 
the sensor and algorithm parameters, and on the environment \cite{Soatto11active}.
A complete treatment of how to choose algorithms and sensors to achieve the best performance has not been found yet.
A preliminary study by Censi et al. \cite{Censi15icra}, has shown that the performance for a given task also depends on the power available for sensing.
It also suggests that the optimal sensing architecture may have multiple sensors that might be instantaneously 
switched on and off according to the required performance level \blue{or measure the same phenomenon through different physical principles for robustness~\cite{durrant1996autonomous}}.

 \subsection{Deep Learning}
 \label{sec:deepLearning}

It would be remiss of a paper that purports to consider future directions in SLAM not to make mention of deep learning.  
Its impact in computer vision has been transformational, and, at the time of writing this article, it is already making significant inroads into traditional robotics, including SLAM.

Researchers have already shown that it is possible to learn a deep neural network to regress the inter-frame pose between two images acquired from a moving robot directly from the original image pair \cite{deepVO-RAL2016}, effectively replacing  the standard geometry of visual odometry.
Likewise it is possible to localize the 6DoF of a camera  \blue{ with regression forest \cite{Valentin2015cvpr} and with} deep convolutional neural network \cite{kendall2015convolutional}, and to estimate the depth of a scene (in effect, the map) from a single view solely as a function of the input image \cite{eigen-iccv2015,cadenaMAE-RSS16,liu-cvpr2015}.

This does not, in our view, mean that traditional SLAM is dead, and it is too soon to say whether these methods are simply curiosities that show what can be done in principle, but which will not replace traditional, well-understood methods, or if they will completely take over. 

\myParagraph{Open Problems} We highlight here a set of future directions for SLAM where we believe
\blue{
machine learning and more specifically deep learning will be influential, or where the SLAM application will throw up challenges for deep learning.
}

\mySubParagraph{Perceptual tool}
It is clear that some perceptual problems that have been beyond the reach of off-the-shelf computer vision algorithms can now be addressed.
For example, object recognition for the imagenet classes \cite{imagenet2015} can now, to an extent, be treated as a black box that works well from the perspective of the roboticist or SLAM researcher.
Likewise semantic labeling of pixels in a variety of scene types reaches performance levels of around 80\% accuracy or more \cite{everingham2010pascal}.
We have already commented extensively on a move towards more semantically meaningful maps for SLAM systems, and these black-box tools will hasten that.
But there is more at stake:  deep networks show more promise for connecting raw sensor data to understanding, \blue{or connecting raw sensor data to actions,} than anything that has preceded them.

\mySubParagraph{Practical deployment}
Successes in deep learning have mostly revolved around lengthy training times on supercomputers and 
inference on special-purpose GPU hardware for a one-off result.
A challenge for SLAM researchers (or indeed anyone who wants to embed the impressive results in their system) is how to provide sufficient computing power in an embedded system.
Do we simply wait for the technology to catch up, or do we investigate smaller, cheaper networks that can produce ``good enough'' results, and consider the impact of sensing over an extended period?

\blue{
\mySubParagraph{Online and life-long learning}
An even greater and important challenge is that of online learning and adaptation, that will be essential to any future long-term SLAM system.  SLAM systems typically operate in an open-world with continuous observation, where new objects and scenes can be encountered.  But to date deep networks are usually trained on closed-world scenarios with, say, a fixed number of object classes. A significant challenge is to harness the power of deep networks in a one-shot or zero-shot scenario (i.e. one or even zero training examples of a new class) to enable life-long learning for a continuously moving, continuously observing SLAM system.

Similarly, existing networks tend to be trained on a vast corpus of labelled data, yet it cannot always be guaranteed that a suitable dataset exists or is practical to label for the supervised training.  One area where some progress has recently been made is that of single-view depth estimation: Garg\setal~\cite{Garg:etal:ECCV2016} have recently shown how a deep network for single-view depth estimation can be trained simply by observing a large corpus of stereo pairs, without the need to observe or calculate depth explicitly.  It remains to be seen if similar methods can be developed for tasks such as semantic scene labelling.   
}

\mySubParagraph{Bootstrapping}
Prior information about a scene has increasingly been shown to provide a significant boost to SLAM systems.
Examples in the literature to date include known objects \cite{SalasMoreno13cvpr,dame-cvpr2013} or prior knowledge about the expected structure in the scene, like smoothness as in DTAM \cite{Newcombe11iccv}, Manhattan constraints as in \cite{flint2011manhattan}, or even the expected relationships between objects \cite{bao-cvpr12}.
It is clear that deep learning is capable of distilling such prior knowledge for specific tasks such as estimating scene labels or scene depths.
How best to extract and use this information is a significant open problem.  \blue{It is more pertinent in SLAM than in some other fields because in SLAM we have solid grasp of the mathematics of the scene geometry -- the question then is how to fuse this well-understood geometry with the outputs of a deep network.  One particular challenge that must be solved is to characterize the uncertainty of estimates derived from a deep network.}

SLAM offers a challenging context for exploring potential connections between deep learning architectures and recursive state estimation in large-scale graphical models.
For example, Krishan\setal~\cite{krishnan2015deep} have recently proposed Deep Kalman Filters; perhaps it might one day be possible to create an end-to-end SLAM system using a deep architecture, without explicit feature modeling, data association, etc.




\section{Conclusion}
\label{sec:conclusion}

The problem of simultaneous localization and mapping has seen great
progress over the last 30 years.  Along the way, several important
questions have been answered, while many new and interesting questions
have been raised, with the development of new applications, new
sensors, and new computational tools.

\blue{Revisiting the question ``is SLAM necessary?'', we believe the
  answer depends on the application, but quite often the answer is a
  resounding \emph{yes}.} \blue{SLAM and related techniques, such as
 visual-inertial odometry, are being increasingly
  deployed in a variety of real-world settings, from self-driving cars
  to mobile devices.  SLAM techniques will be increasingly relied upon
  to provide reliable metric positioning in situations where
  infrastructure-based solutions such as GPS are unavailable or do not
  provide sufficient accuracy.  One can envision cloud-based
  location-as-a-service capabilities coming online, and maps becoming
  commoditized, due to the value of positioning information for mobile
  devices and agents.

In some applications, such as self-driving cars, precision
localization is often performed by matching current sensor data to a
high definition map of the environment that is created in
advance~\cite{Levinson11thesis}.  If the a priori map is accurate,
then online SLAM is not required.  Operations in highly dynamic
environments, however, will require dynamic online map updates to deal
with construction or major changes to road infrastructure. The
distributed updating and maintenance of visual maps created by large
fleets of autonomous vehicles is a compelling area for future work.

One can identify tasks for which different flavors of SLAM
formulations are more suitable than others.  For instance, a
topological map can be used to analyze reachability of a given place,
but it is not suitable for motion planning and low-level control; a
locally-consistent metric map is well-suited for obstacle avoidance
and local interactions with the environment, but it may sacrifice
accuracy; a globally-consistent metric map allows the robot to perform
global path planning, but it may be computationally demanding to
compute and maintain.

One may even devise examples in which SLAM is unnecessary altogether
and can be replaced by other techniques, e.g., visual servoing for
local control and stabilization, or ``teach and repeat'' to perform
repetitive navigation tasks.  A more general way to choose the most
appropriate SLAM system is to think about SLAM as a mechanism to
compute a sufficient statistic that summarizes all past observations
of the robot, and in this sense, which information to retain in this
compressed representation is deeply task-dependent.}

\blue{
As to the familiar question ``is SLAM solved?'', in this position paper we argue that, as we enter the \emph{robust-perception age}, the question cannot be answered without specifying a robot/environment/performance combination.
For many applications and environments, numerous major challenges and important questions remain open.  
To achieve truly robust perception and navigation for long-lived autonomous robots, more research in SLAM is needed.
As an academic endeavor with important real-world implications, SLAM is \emph{not} solved.
}

The unsolved questions involve four main aspects: \emph{robust performance}, \emph{high-level understanding}, \emph{resource awareness}, and \emph{task-driven inference}. 
From the perspective of robustness, the design of fail-safe, self-tuning SLAM systems is a
formidable challenge with many aspects being largely unexplored.
For long-term autonomy, techniques to construct and maintain
large-scale time-varying maps, as well as policies that define when to
remember, update, or forget information, still need a large amount of
fundamental research; similar problems arise, at a different scale, in
severely resource-constrained robotic systems.

Another fundamental question regards the design of metric and semantic
representations for the environment. Despite the fact that the
interaction with the environment is paramount for most applications of robotics, modern SLAM systems are not able to provide a  tightly-coupled high-level understanding of the geometry and the
semantic of the surrounding world; the design of such representations
must be task-driven and currently a tractable framework to link task
to optimal representations is lacking.  \blue{Developing such a
  framework will bring together the robotics and computer vision
  communities.}  

Besides discussing many accomplishments and future challenges for the SLAM community, 
we also examined opportunities connected to the  use 
of novel sensors, new  tools 
(e.g., convex relaxations and duality theory, or deep learning), and the role of active sensing. 
%
SLAM still constitutes an indispensable backbone for most robotics applications and 
despite the amazing progress over the past decades, existing SLAM systems are far from 
providing insightful, actionable, and compact models of the environment, comparable to
the ones effortlessly created and used by humans.


\section*{Acknowledgment}
\label{sec:acknowledgments}

%
%

We would like to thank all the contributors and members of the Google Plus community~\cite{RSS15website}, especially Liam Paul, Ankur Handa, Shane Griffith, Hilario Tom\'e, Andrea Censi, and Raul Mur Artal, for their inputs towards the discussion of the open problems in SLAM.
We are also thankful to all the speakers of the RSS workshop~\cite{RSS15website}:
Mingo Tardos, Shoudong Huang, Ryan Eustice, Patrick Pfaff, Jason Meltzer, Joel Hesch, Esha Nerurkar, Richard Newcombe, Zeeshan Zia, Andreas Geiger.
We are also indebted to Guoquan (Paul) Huang and Udo Frese for the discussions during the preparation of this document,  and Guillermo Gallego and Jeff Delmerico for their early comments on this document.

\bibliographystyle{abbrv}
\footnotesize{
\bibliography{./bibliography/final_acronyms,./bibliography/final_refs}
}




\end{document}